%% file: paper.tex
\documentclass[runningheads]{llncs}

\input{math_commands.tex}

\usepackage[T1]{fontenc}
\usepackage[utf8]{inputenc}
\usepackage{url}
\usepackage{graphicx}
\usepackage[hidelinks]{hyperref}

\usepackage{tikz}
\usetikzlibrary{shapes,shapes.geometric,backgrounds,calc,fit,positioning}
\makeatletter
\tikzset{circle split part fill/.style  args={#1,#2}{%
 alias=tmp@name,
  postaction={%
    insert path={
     \pgfextra{%
     \pgfpointdiff{\pgfpointanchor{\pgf@node@name}{center}}%
                  {\pgfpointanchor{\pgf@node@name}{east}}%
     \pgfmathsetmacro\insiderad{\pgf@x}
      \fill[#1] (\pgf@node@name.base) ([xshift=-\pgflinewidth]\pgf@node@name.east) arc
                          (0:180:\insiderad-\pgflinewidth)--cycle;
      \fill[#2] (\pgf@node@name.base) ([xshift=\pgflinewidth]\pgf@node@name.west)  arc
                           (180:360:\insiderad-\pgflinewidth)--cycle;
         }}}}}
 \makeatother

\usepackage[ruled,vlined]{algorithm2e}
\usepackage{mathtools}
\usepackage{commath}
\usepackage{faktor}
\usepackage{microtype}
\usepackage{nicefrac}
\usepackage{colonequals}
\newcommand*{\longeq}{\ratio\Longleftrightarrow}
\usepackage{enumerate}
\usepackage{paralist}
\usepackage{multirow}
\usepackage[noabbrev]{cleveref}
\let\cref\Cref

\usepackage{booktabs}
\usepackage{arydshln}

\usepackage{xcolor}
\usepackage{colortbl}

\usepackage{fca, newdrawline}

\newcommand{\Scon}{\mathbb{S}}

\begin{document}

\title{Conceptual Views of Neural Networks:\\A Framework for Neuro-Symbolic Analysis}
\titlerunning{Conceptual Views of Neural Networks}

\author{Johannes Hirth\inst{1}\orcidID{0000-0001-9034-0321} \and
  Tom Hanika\inst{2}\orcidID{0000-0002-4918-6374}}

\authorrunning{J. Hirth, T. Hanika}

\institute{ Knowledge \& Data Engineering, University of Kassel, Germany\\
\and
 Information Systems and Machine Learning Lab, University of Hildesheim, Germany\\
\email{hirth@cs.uni-kassel.de, hanika@uni-hildesheim.de}}

\maketitle

\begin{abstract}
  We introduce \emph{conceptual views} as a formal framework grounded
  in Formal Concept Analysis for globally explaining neural
  networks. Experiments on twenty-four ImageNet models and Fruits-360
  show that these views faithfully represent the original models,
  enable architecture comparison via Gromov--Wasserstein distance, and
  support abductive learning of human-comprehensible rules from
  neurons.
\end{abstract}

\section{Introduction}\label{sec:intro}

Neural networks (NNs) are known for their strong performance across a
wide range of learning problems. However, these results are almost
always achieved at the cost of human
explainability. This tension is addressed from different standpoints in
the literature. On the one hand, there are calls to refrain from using
NNs for high-stakes decisions and to rely on inherently interpretable
methods, even at the cost of
accuracy~\cite{rudin2019explaining}. On the other hand, a substantial
body of research aims to develop methods for explaining NN
models. Such explanations can be broadly classified as \emph{local
  explanations}, which address why a particular input was treated in a
specific way~\cite{lime}, and \emph{global explanations}, which seek
to characterize the model as a whole. While local methods such as
saliency maps or activation highlighting~\cite{net2vec} are effective
for flat data like images, they are less applicable to
high-dimensional or complex data where human visual inspection is
infeasible. Global approaches, by contrast, are inherently more
challenging and therefore less explored, yet they are essential for
comprehensive understanding of NNs.

In this work, we propose \emph{conceptual views} as a formal
framework for the global analysis of neural networks.  Our approach
targets the embedding space of a neural network, specifically the
outputs at the last hidden layer. We introduce two complementary
representations:
\begin{inparaenum}
\item A \emph{many-valued conceptual view} that captures the
  real-valued activation and weight structure as a pair of matrices,
  inducing a pseudo-metric space on objects and classes.
\item A \emph{symbolic conceptual view} that discretizes this
  structure via conceptual scaling~\cite{scaling} from Formal Concept
  Analysis (FCA)~\cite{fca-book}, yielding a binary relational
  representation.
\end{inparaenum}

The key advantage of our framework is its composability: multiple
neuro-symbolic explanation procedures can be integrated, their results
merged within a concept lattice structure, and additional parts of the
representation decoded using background knowledge. The concept lattice
further enables various forms of symbolic reasoning, including the
extraction of propositional implications.

We demonstrate through extensive experiments that: (i) many-valued
conceptual views achieve high fidelity as surrogates for the original
models; (ii) the Gromov-Wasserstein distance applied to conceptual
views yields meaningful similarity measures between architectures;
(iii) symbolic conceptual views, particularly with Tanh activation,
enable decision-tree based surrogates with competitive fidelity; and
(iv) background knowledge can be integrated via subgroup discovery to
derive human-comprehensible rules about neurons.

\section{Related Work}\label{sec:related}

\textbf{Local and Concept-Based Explanations.}
Several approaches aim to provide insights into neural networks by
highlighting parts of the input relevant to a particular
prediction~\cite{lime}, so-called \emph{local explanations}. However,
these rely on the user's ability to comprehend input
representations. To overcome this limitation, the state of the art
employs \emph{symbolic concepts} as an intermediate
representation~\cite{NeSyAI}. Methods such as
Net2Vec~\cite{net2vec} classify network activations into pre-defined
concepts but require manually crafted input representations.
Particularly influential is \emph{TCAV}~\cite{tcav}, which quantifies
the importance of user-defined concepts via directional derivatives,
complemented by \emph{ACE}~\cite{ACE}, which automatically identifies
coherent activation patterns at a given layer.  A comprehensive survey
is provided by Lee et~al.~\cite{concept-survey-lee2025}.

\textbf{Concept Bottleneck Models.}
A significant recent development in concept-based explainability is
the \emph{Concept Bottleneck Model} (CBM)~\cite{cbm-koh2020}, which
constrains the network to first predict human-interpretable concepts
before mapping them to task labels. This paradigm has been extended by
post-hoc methods~\cite{posthoc-cbm-yuksekgonul2023} that retrofit
concept bottlenecks onto pre-trained models, and by label-free
approaches~\cite{labelfree-cbm-oikarinen2023} that leverage
vision-language models to discover concepts without manual annotation.
More recently, Mangal et~al.~\cite{concept-vlm-mangal2024} leverage
vision-language models to map internal representations to
human-understandable concepts, enabling specification-level
verification. While CBMs enforce concept-level interpretability by
architectural design, our framework takes a complementary approach:
rather than constraining the architecture, we analyze the learned
representations post-hoc using formal algebraic structures.

\textbf{Neural Network Similarity.}
Comparing what different neural networks have learned is a question of
growing importance, especially in the context of transfer learning and
model selection. Centered Kernel Alignment
(CKA)~\cite{cka-kornblith2019} provides a similarity measure between
representations of different networks by comparing their activation
patterns. Our approach differs in that we do not compare activations
directly but instead operate on the pseudo-metric spaces induced by
conceptual views, employing the
Gromov-Wasserstein~\cite{GW-dist} distance. This is closely related
to the idea of relating neural networks to kernel
spaces~\cite{pmlr-v119-shankar20a,NEURIPS2019_0d1a9651}, but our
framework investigates the space itself rather than the mapping into it.

\textbf{Mechanistic Interpretability.}  Recent advances in mechanistic
interpretability~\cite{mechinterp-olah2020,mechinterp-conmy2023},
surveyed comprehensively by Bereska and
Gavves~\cite{mechinterp-bereska2024}, aim to reverse-engineer neural
network computations at the level of individual neurons and
circuits. While mechanistic approaches provide fine-grained insights
into how specific computations are implemented, they typically operate
at a local or circuit level. Our framework complements this by
providing a global perspective on the NN's learned knowledge, grounded
in the well-established theory of FCA.

\textbf{Formal Concept Analysis in Machine Learning.}
FCA~\cite{fca-book,Wille1982} provides a mathematically principled
framework for deriving conceptual hierarchies from binary data, with a
long history in knowledge representation. Its application to machine
learning has been explored for decision tree
induction~\cite{conf/cla/BelohlavekBOV07} and classification.
The conceptual scaling theory~\cite{scaling} provides a
well-elaborated tool-set for translating many-valued data into formal
contexts. Our prior work on scale-measures~\cite{smeasure} and
conceptual error quantification~\cite{sm-error} provides the
theoretical basis for the symbolic translation employed here.

\textbf{Neuro-Symbolic AI.}
The broader field of neuro-symbolic AI~\cite{NeSyAI} seeks to combine
the learning capabilities of neural networks with the reasoning
capabilities of symbolic systems. Our work contributes to this
direction by providing a principled bridge: the symbolic conceptual
view translates neural representations into formal contexts, which can
then serve as input for classical symbolic reasoning methods such as
description logics and subgroup discovery.

\section{The Views of Neural Networks}\label{sec:mv-views}

To extract the inherent knowledge captured by a neural network, we
propose a two-staged approach.
First, we derive a formal relational structure that describes how the
data is represented by the last hidden layer of a neural network. For
this, we employ conceptual scaling from Formal Concept Analysis, which
provides a formal and comprehensible method to transform data into a
symbolic space. The derived symbolic attributes are user defined and
depend on the data. A key idea here is that these attributes are
simple, allowing the analyst to relate symbolic deductions to
subspaces of the real-valued data. Throughout the remainder of this
section, we propose the use of thresholds $\delta$ for individual
neuron activations. This introduces two symbolic attributes
$n_{\leq \delta}$ and $n_{\geq \delta}$ that are in relation to
entities that have an activation above or below that threshold. While
this process is intentionally straightforward, we show in the
following section empirically that the choice of $\delta$ is a
difficult problem that depends on the learning problem.  We illustrate
this approach in \cref{fig:views}.
In the second step, we decode the symbolic space using background
knowledge and symbolic features derived from concept-based
explanations. We elaborate on this in greater detail in
\cref{sec:abductive}.

\begin{figure*}[ht]
  \centering
  \resizebox{0.9\textwidth}{!}{
    \input{pics/pipline.tikz}
  }

  \caption{A simplified neural network drawing (left), its
    many-valued conceptual view (middle), and its
     symbolic conceptual view (right).}
  \label{fig:views}
\end{figure*}

\subsection{Many-Valued Conceptual View}\label{sec:mv-view-def}
We introduce two notions of conceptual view of a neural network: a
\emph{many-valued} and a \emph{symbolic} view. Both provide novel
methods to enable a human AI analyst to grasp deeper insights into the
knowledge captured by the neurons. In addition, the symbolic view
facilitates the application of abductive learning procedures. This
results in rules that allow explaining a NN by means of
human-comprehensible terminology, as well as in terms of the neurons.

Let $N$ be the set of neurons of the last hidden layer of a NN. We
interpret NNs as a map from input objects $g\in G$,
represented as $g=(v_1,\dots,v_m)\in \mathbb{R}^{m}$, to outputs in
$[0,1]^{\abs{C}}$ for classes $C$. The parameter $m$ specifies the
number of input features (see \cref{fig:views}). Naturally, we 
interpret each $n\in N$ as a function by itself from the input
layer up to the activation of $n$, i.e.,
$n:\mathbb{R}^{m} \to \mathbb{R}$.  The output neurons can be
characterized analogously by
$c:\mathbb{R}^{\abs{N}} \to \mathbb{R}$.  With $w_{i,j}$ we address
the weights connecting the output neuron $c_i\in C$ with hidden neuron
$n_j\in N$.

\begin{definition}[Many-Valued Conceptual View]\label{def:view}
  For a neural network NN, $C$ its output classes and
  $N=\{n_1,\dots,n_h\}$ the neurons of the last hidden layer.  We
  define the \emph{many-valued conceptual view} as
  $\mathcal{V}=(\mathbb{O},\mathbb{W})$, where
  $\mathbb{O}\in \mathbb{R}^{\abs{G}\times \abs{N}}$ with value at
  $(i,j)$ equal to the activation $n_j(g_i)$, called \emph{Object
    View}, and $\mathbb{W}\in \mathbb{R}^{\abs{C}\times \abs{N}}$ with value at
  $(i,j)$ equal to the weight $w_{i,j}$, called \emph{Class View}.
\end{definition}

A short motivation: with the object view $\mathbb{O}$, we study the
activation of the neurons $N$ given an object $g$.  Complementarily,
with the class view $\mathbb{W}$, we investigate the relation of the
neurons $N$ to the outputs $c\in C$ by their corresponding weights
$w_{i,j}$. For example, given~\cref{fig:views}, which depicts the
object and class view (right) of the network (left). We find that
$n_k(o_t)$ is greater than $n_1(o_t)$, from which we infer that the
relation of $o_t$ to $n_k$ is greater than $n_1$.

We employ the introduced views to comprehend the complete
classification captured by a NN model. We can represent any object $g$
as a row in the object view matrix, i.e.,
$O(g)\coloneqq(n_1(g),\dots,n_h(g))$. Analogously, we can represent
any class $c_i$ as a row in the class view matrix, i.e.,
$W(c_{i})\coloneqq(w_{i,1},\dots,w_{i,h})$. The outputs of the NN for
class $c_i$ follow from the term $O(g)\cdot W(c_i) + b$, where $b$ is
a bias. This can be rewritten as
$\abs{O(g)}\cdot \abs{W(c{_i})}\cos(O(g),W(c_{i})) + b$ where
$\cos(O(g),W(c_{i}))$ is the cosine value of the angle between $O(g)$
and $W(c_{i})$.

Thus, to understand the inner representation of the classes $C$ within
the NN, it is reasonable to grasp the objects and classes in the same
space and classify objects using similarity measures. Using this we
can introduce an \emph{object-class distance map}
$d_{\mathcal{V}}:G\times C \to \mathbb{R}, (g,c)\mapsto d(O(g),W(c))$,
where a sensible choice for $d$ is \emph{cosine similarity} or the
\emph{Euclidean distance}. We investigate both in
\cref{sec:eval-mv-view}.

Hence, using $d_{\mathcal{V}}$ and similar distance maps for
$G\times G$ and $C\times C$, one can derive a pseudo-metric space
$(G\cup C,\hat{d}_{\mathcal{V}})$. From this representation of $G$
and $C$ one can infer a simple classification map, e.g., by applying
1-NN classification.

\textbf{Similarity of Neural Networks.}
Conceptual views enable a direct comparison of NNs. We employ the
Gromov-Wasserstein (GW) distance~\cite{GW-dist}, as experimentally
demonstrated in~\cref{sec:sim-exp}. The GW distance solves an
underlying optimization problem that seeks isometric mappings of the
pseudo-metric spaces of two neural networks into a common third space,
minimizing the deviation of object representations. In contrast to the
Gromov-Hausdorff distance, the GW distance allows for matchings
instead of one-to-one mappings between objects, making it
computationally  tractable.
We note two important observations. First, our approach for similarity
is related~\emph{kernel spaces}~\cite{pmlr-v119-shankar20a,NEURIPS2019_0d1a9651},
which enables studying how objects are \emph{hierarchically clustered}
in such a space. Our notion, however, does not consider how objects are
mapped into this space but rather investigates the space itself.
Second, the GW distance is invariant with respect to permutations of
the many-valued conceptual views.

\subsection{Symbolic View}\label{sec:s-views}
We employ \emph{conceptual~scaling}~\cite{scaling}
from formal~concept~analysis (FCA), where data is represented in a
\emph{formal~context} $\context=(G,M,I)$. There, $G$ is a set of
objects, $M$ a set of attributes and $I\subseteq G\times M$ is an
incidence relation, where $(g,m)\in I$ indicates that $g$ has
attribute $m$. From $I$ arise two \emph{derivation operators}
$(\cdot)^{I}:\mathcal{P}(G)\to \mathcal{P}(M)$, with $A^{I}=\{m\in
M\mid \forall g\in A:(g,m)\in I\}$, and analogously
$(\cdot)^{I}:\mathcal{P}(M)\to \mathcal{P}(G)$, with $B^{I}=\{g\in
G\mid \forall m\in B:(g,m)\in I\}$.
Applying many-valued conceptual scaling to a many-valued data set
yields a formal context. We employ \emph{dichotomic scaling}, using
thresholds for the object view $\delta_{\mathbb{O}}$ and the class
view $\delta_{\mathbb{W}}$ (see \cref{fig:views}). The resulting
relational structure is invariant with respect to row- or column
permutations in the related many-valued conceptual view
(\cref{def:view}).

\begin{definition}[Symbolic Conceptual View]
  Let $\mathcal{V}=(\mathbb{O},\mathbb{W})$ the many-valued conceptual
  view of a NN and let $\delta_{\mathbb{O}},\delta_{\mathbb{W}}$ be
  threshold values. We define the \emph{symbolic conceptual view}
  $\mathcal{V}_{\mathbb{D}}=(\mathbb{O}_{\mathbb{D}},\mathbb{W}_{\mathbb{D}})$ by
  \begin{align}
    \mathbb{O}_{\mathbb{D}}\coloneqq
    (G,N\cup \bar{N},I_{\mathbb{O}}),\text{ with }&\begin{aligned}[t]&(g,n_j)\in I_{\mathbb{O}} \longeq
    n_j(g)> \delta_{\mathbb{O}}  \text{ and }\\ &(g,\bar{n}_j)\in I_{\mathbb{O}} \longeq
    n_j(g)\leq \delta_{\mathbb{O}}\end{aligned} \tag{Symbolic Object View}\\
    \mathbb{W}_{\mathbb{D}}\coloneqq
    (C,N\cup \bar{N},I_{\mathbb{W}}),\text{ with }& \begin{aligned}[t]& (c_i,n_j)\in I_{\mathbb{W}} \longeq
    w_{i,j}> \delta_{\mathbb{W}} \text{ and }\\
                                                                      & (c_i,\bar{n}_j)\in I_{\mathbb{W}} \longeq w_{i,j}\leq \delta_{\mathbb{W}}. \end{aligned}\tag{Symbolic Class View}
  \end{align}
with artificial symbols  $\bar{N}\coloneqq \{\bar{n}\mid n\in N\}$  used  as defined above.
\end{definition}

This definition allows for constructing human-comprehensible
explanations given a background ontology, e.g., from human
annotations, ontological databases, or from concept-based
explanations. We exemplify this in \cref{fig:views} using the formal
context $\Scon_{N}$ that employs interpretable features
$S_{m_1},\dotsc, S_{m_l}$. We provide more details in
\cref{sec:abductive}. Suitable threshold values
$\delta_{\mathbb{W}},\delta_{\mathbb{O}}$ depend on the architecture
of the NN model under analysis. For example, if the activation
function is ReLU, the neuron's co-domain is positive. Thus it becomes
difficult to determine a reasonable $\delta$ for negative symbols
$\bar{N}$, as studied in~\cref{sec:symb-con-view}.

\textbf{Evaluating Symbolic Conceptual Views}

Crucial for the quality of the derived symbolic representations are
two criteria. First, classes need to be uniquely represented in
$\context[W]_{\context{D}}$. Second, the symbolic features should
allow for straightforward descriptions for each class in $C$. This can
be measured by the performance of interpretable classifiers such as
decision trees or nearest-neighbor classifiers.

\section{Experimental Study}\label{sec:experiments}

We support our formal framework of conceptual views through an
experimental study using common and well-known data sets and NN
models. Our experimental design follows a principled evaluation
strategy where each transformation step (from the original model
through the many-valued view to the symbolic view) is separately
validated. First, we evaluate the fidelity of the many-valued
conceptual view by a classification task
(\cref{sec:eval-mv-view}). Second, we show how to compare many-valued
conceptual views of different NN models
(\cref{sec:sim-exp}). Third, we derive human-comprehensible
representations via symbolic views, conducting a parameter study to
identify reasonable threshold values before validating them via
classification (\cref{sec:symb-con-view}). The code and trained
models are publicly
available.\footnote{\url{https://github.com/FCA-Research/Formal-Conceptual-Views-in-Neural-Networks}}

Although the scaling procedure is conceptually simple, it is not
trivially effective. The quality of the resulting views depends on the
activation function, the distribution of activations and weights, and
the number of neurons. Each step must therefore be carefully
evaluated, which explains the staged design.

\subsection{Many-Valued Conceptual Views on ImageNet}\label{sec:eval-mv-view}
We demonstrate that many-valued conceptual views are capable of
capturing a large share of a NN model's behavior. For this, we use all
twenty-four NN models from TensorFlow that are trained on the
ImageNet~\cite{imagenet} data set (as of July 2022). The object view
is calculated using the test set, i.e., 100k images, of
\emph{ImageNet} used in the ILSVRC~\cite{imagenet-ILSVRC}
challenge. In the Appendix\footnote{The appendix can be found in the arXiv version~\url{https://arxiv.org/abs/2209.13517v2}.}, \cref{tab:weights-long}, we compile basic
statistics on these networks and our views.
The models span a variety of architectures and employ either ReLU or
Swish activation functions in the last hidden layer. The number of
neurons ranges from 1,024 (MobileNet) to 4,096 (VGG). Weight means
are near zero across all models, while activation distributions vary
considerably depending on the architecture (see
Appendix,~\cref{tab:weights-long}).

\textbf{Evaluating ImageNet Views.}
To evaluate the quality of our views, we compare a
one-nearest-neighbor (1-NN) classifier on the pseudo-metric space
$(G\cup C,\hat{d}_{\mathcal{V}})$ introduced in \cref{sec:mv-views}
directly with the NN classification function on all 100,000 test
images. We use model fidelity as the evaluation criterion, i.e., we
count instances where the 1-NN outputs the same class label as the NN
and normalize by the cardinality of the test set. We differentiate
between using cosine similarity and Euclidean distance within
$\hat{d}_{\mathcal{V}}$.
\begin{table}[b]
  \caption{Fidelity between a NN model and its many-valued conceptual
    view using 1-NN. All twenty-four results are in
    Appendix,~\cref{tab:imagenet-fidel-long}.}
  \centering
  \setlength{\tabcolsep}{3pt}
    \begin{tabular}{|l||c|c|c|c|c|c|}
      \hline
      &VGG16&DenseNet201&MobilNetV1&ResNet152V2&EffB0&EffB7\\ \hline \hline
      Euclidean&0.945&0.972&0.575&0.999&0.944&0.985\\
      Cosine&0.841&0.728&0.449&0.314&0.933&0.979\\
      \hline
    \end{tabular}
  \label{tab:imagenet-fidel}
\end{table}

\textbf{Observations.}
We find that the many-valued conceptual view achieves high fidelity
across the majority of models (see \cref{tab:imagenet-fidel}). The
MobileNetV1 model is the only notable exception, which we attribute to
its compact architecture and aggressive dimensionality reduction.
Euclidean distance is consistently superior to cosine similarity,
with particularly large gaps for ResNet models (up to 0.6
difference). For the EfficientNet family, we observe a near-monotone
relationship between the number of neurons $|N|$ in the last hidden
layer and fidelity. These results suggest that the many-valued
conceptual view is a meaningful representation and that classification
functions based on the resulting pseudo-metric space can serve as
faithful surrogates for the NN model.

We acknowledge that the choice of distance metric is a design
decision that may not generalize uniformly across all architectures.
However, the consistently strong performance of Euclidean distance
provides a practical default. Future work could investigate
learned distance metrics  to further
improve fidelity, particularly for compact architectures where the
simple metrics show limitations.

\subsection{Similarity of Neural Networks}\label{sec:sim-exp}
\begin{figure}[t]
  \centering
    \includegraphics[width=0.4\textwidth, trim=0 5 0 7, clip]{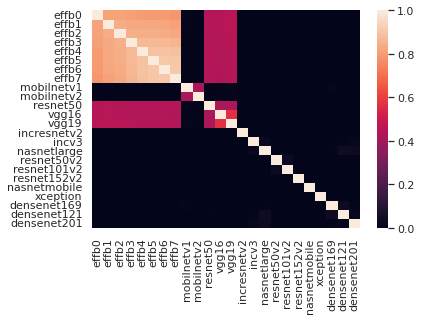}
  \includegraphics[width=0.4\textwidth, trim=0 5 0 7, clip]{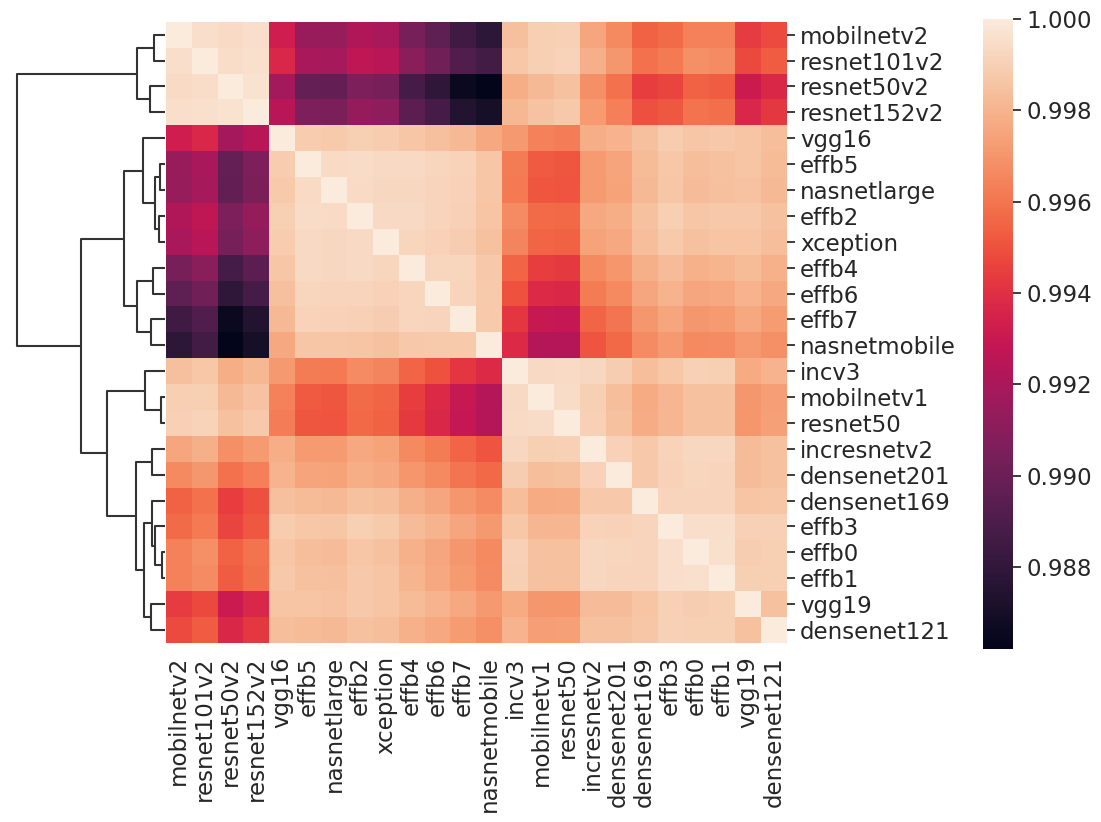 }
  \caption{Similarity of twenty-four neural networks trained on
    ImageNet. Left: pairwise fidelity baseline. Right: similarity
    using Gromov-Wasserstein distance on the class view. The object
    view is in Appendix,~\cref{fig:imagenet-sim-long}.}
  \label{fig:imagenet-sim}
\end{figure}

From many-valued conceptual view we derive for all NN models a
pseudo-metric space as introduced in \cref{sec:mv-views}. Using
the GW distance~\cite{GW-dist} to compare these spaces. In
\cref{fig:imagenet-sim} (right) we depict the individual distances for
all considered models with respect to the class and object view. We
used ten percent of the test data set and applied a uniform
probability measure on the data points. This is compared to a
baseline derived from pairwise fidelity (\cref{fig:imagenet-sim},
left).

\textbf{Observations.}
From the pairwise fidelity diagram (\cref{fig:imagenet-sim}, left), we
infer that almost all models are distinct with the exception of VGG16,
VGG19, ResNet50, and the EfficientNet instances, where later models
become more similar with increasing neuron count. The GW-based
similarity plots reveal a different and richer structure: we can
visually identify clusters of models that often correspond to similar
architectures. In the class view, this clustering is finer than in the
object view, suggesting that the weight structure captures
architecture-specific patterns more effectively.
A natural question is whether similarity measures such as CKA~\cite{cka-kornblith2019}
would yield comparable clusterings. While CKA operates directly on
activation matrices and is well-suited for layer-wise comparisons, our
GW-based approach operates on the full pseudo-metric space and
is invariant under permutations of neurons, a property that is
important when comparing architectures with different internal orderings.
A detailed comparative study between these approaches is an
interesting direction for future work that would, however, exceed the
scope of this paper due to the page limit.

\subsection{Symbolic Conceptual View}
\label{sec:symb-con-view}
We study the influence of activation functions and the number of
neurons on the determination of threshold
values for meaningful symbolic conceptual views.

\textbf{Parameter Study.} To evaluate the influence of the activation
function and the number of neurons, we trained one NN architecture
multiple times on the \emph{Fruits-360}~\cite{fruit360} data set,
which contains 67,692 images of 131 types of fruits or vegetables
(plus 22,688 test images).  We used the architecture from the
Fruits-360
experiment\footnote{\url{https://github.com/Horea94/Fruit-Images-Dataset}}
with parameters from~\cite{fruit360}, modifying the last two hidden
layers. For the last hidden layer $N$ we vary its size $2^{n}$ between
$2^{4}$ and $2^{9}$. For the penultimate layer we use $2^{\lfloor
  \frac{10+n}{2} \rfloor}$ for dimensional reduction. We
studied the impact of \emph{ReLU}, \emph{Linear}, \emph{Swish}, and
\emph{Tanh} activation functions across all layers.
%
%
%
For each parameter setting we trained ten models and computed their
respective conceptual views.

\begin{figure}[b!]
  \centering
  \begin{tabular}{lm{0.85\linewidth}}
    $\mathbb{O}$& \includegraphics[width=0.9\linewidth, trim=0 8 0 8, clip]{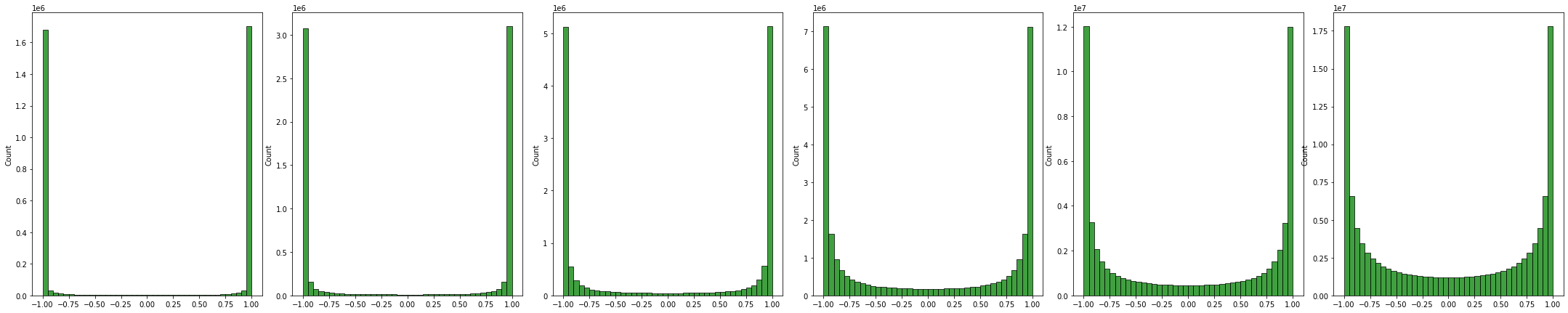}\\
    $\mathbb{W}$& \includegraphics[width=0.9\linewidth, trim=0 9 0 16, clip]{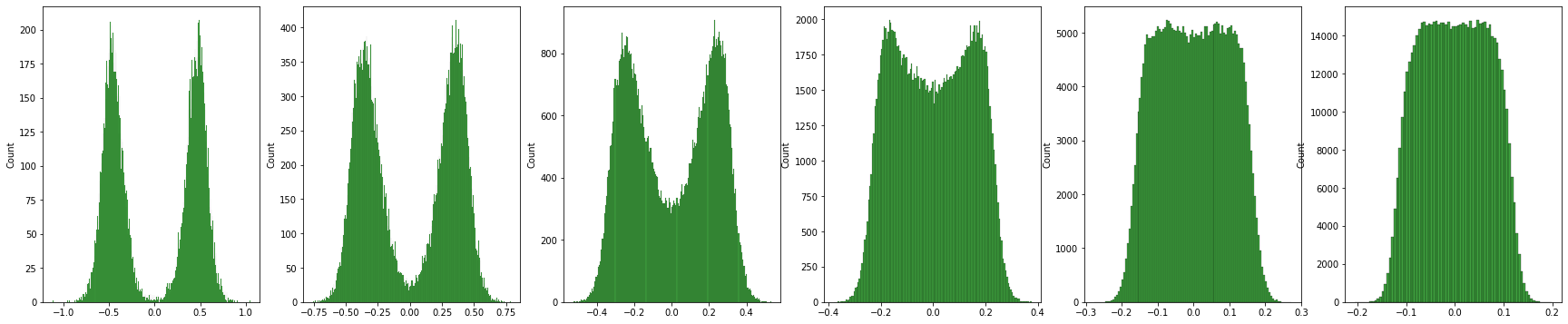}
  \end{tabular}
  \caption{Value distributions for the object ($\mathbb{O}$) and class
    ($\mathbb{W}$) view for ten runs using Fruits-360 and \textbf{Tanh}
    activation. Last hidden layer sizes from $2^{4}$ (first col.) to
    $2^{9}$ (last col.). Note that the absolute values are not important;
    the relevant aspect is the distributional behavior shown in the plots.
    All activation functions are in
    Appendix,~\cref{fig:weights-long}.}
  \label{fig:weights}
\end{figure}

\begin{table}[t]
  \caption{Parameter study results for Tanh: influence of activation
    function and neuron count on view quality.  V-Fid and SV-Fid
    denote the fidelity of 1-NN to the original model for the
    many-valued and symbolic view, respectively. Full results in
    Appendix,~\cref{tab:distro-stats-long}.}
\centering
\setlength{\tabcolsep}{3pt}
\begin{tabular}{|l||c|c|c|c|c|c|}
    \hline
    &$2^{4}$& $2^{5}$&$2^{6}$&$2^{7}$&$2^{8}$&$2^{9}$\\ \hline\hline
    \multicolumn{7}{|l|}{\textbf{Tanh}}\\\hline
       $\delta_{\mathbb{O}}=0$ Split                             &49.7/50.3  &49.7/50.3  &49.8/50.2  &49.9/50.1  &50.0/50.0  &49.9/50.1\\
       $\delta_{\mathbb{W}}=0$ Split                             &49.9/50.1  &49.8/50.2  &49.8/50.2  &50.0/50.0  &49.9/50.1  &50.0/50.0\\
                       Model Acc                                &90.5$\pm$ 0.8&94.3$\pm$ 0.5&94.7$\pm$ 0.5&94.9$\pm$ 0.4&95.0$\pm$ 0.4&94.8$\pm$ 0.3\\
    V-Fid                                                       &98.3$\pm$ 0.5&99.5$\pm$ 0.1&99.7$\pm$ 0.0&99.7$\pm$ 0.0&99.8$\pm$ 0.0&99.8$\pm$ 0.0\\
    SV-Fid                                                      &94.3$\pm$ 1.4&97.4$\pm$ 0.4&97.7$\pm$ 0.4&97.6$\pm$ 0.1&97.8$\pm$ 0.2&97.6$\pm$ 0.2\\ \hline

  \end{tabular}
  \label{tab:distro-stats}
\end{table}
\textbf{Observations.} The distributions for the object views differ
substantially among activation functions. We found that Tanh produces
the most notable separation of positive and negative values
(see~\cref{fig:weights}). Splitting with $\delta_{\mathbb{O}}=0$ is
meaningful for all examples with respect to separation and symmetry,
yielding two sets of almost equal size. The same holds for
$\delta_{\mathbb{W}}$. We experimented with alternative threshold
determination methods, including mean values, median values,
per-neuron medians, and kernel-density estimation for bivariate
Gaussians, but the split at zero was consistently favorable with
respect to model fidelity.  We report scores for Tanh in
\cref{tab:distro-stats} and for all activation functions in
Appendix,~\cref{tab:distro-stats-long}.

We conclude from our parameter study that Tanh is the recommended
activation function, with
$\delta_{\mathbb{O}},\delta_{\mathbb{W}}=0$.

\textbf{Symbolic Conceptual Views of ImageNet.}
Based on the parameters identified above, we computed the symbolic
conceptual views for the ImageNet models
(results in~\cref{tab:imagenet-fidel-symbolic}).  We found that the
classes are uniquely represented (class separation equals 1), meaning
a perfect classification procedure is theoretically possible using the
symbolic view. However, a direct application of 1-NN using the binary
vectors from the symbolic view yields very poor classification
performance for ReLU-based models. In contrast, the Swish-based models
achieve mediocre to good results, with performance improving for
larger architectures. A reason for the unfavorable ReLU results is its
positive co-domain, which hinders the construction of meaningful
negated attributes $\bar N$. Selected distributions can be found in
Appendix,~\cref{fig:imagenet-distros-long}.
This finding has practical implications: when the goal is to obtain a
faithful symbolic view, one should either use architectures with
symmetric activation functions (such as Tanh) or consider
architecture-specific scaling strategies beyond simple dichotomic
scaling.

\begin{table}[t]
  \caption{Fidelity between six NN models and their symbolic
    conceptual view using 1-NN. All values are in
    Appendix,~\cref{tab:imagenet-fidel-symbolic-long}.}
  \centering
  \setlength{\tabcolsep}{3pt}
    \begin{tabular}{|l||c|c|c|c|c|c|}
      \hline
      &VGG16&DenseNet201&MobilNetV1&ResNet152V2&EffB0&EffB7\\ \hline \hline
      Euclidean&0.552&0.000&0.036&0.000&0.758&0.957\\
      Cos&0.552&0.000&0.036&0.000&0.758&0.957\\
      Class Sep&1.0&1.0&1.0&1.0&1.0&1.0\\
      Activation&ReLU&ReLU&ReLU&ReLU&Swish&Swish\\
      \hline
    \end{tabular}
  \label{tab:imagenet-fidel-symbolic}
\end{table}

\textbf{Symbolic Conceptual Views on Fruits-360.}
The twenty-four ImageNet models employ only ReLU and Swish activation
functions. To complement our study with Tanh, we trained five models
on the Fruits-360 data set: the baseline model
from~\cite{fruit360}, VGG16, ResNet50, IncV3, and EffB0 (the latter
initialized with ImageNet weights). With exception of the baseline, we
added three dense layers (with dropout $p=0.2$) of sizes 1024,
256, and 32, plus an additional layer of size 16 to enable human
explainability through a small neuron count. All added layers employ
Tanh activation; the output layer uses softmax without bias. We used
sparse categorical crossentropy as the loss function.

The results in~\cref{tab:fruit-fidel-symbolic} show that all models
achieve high accuracy, with the four transfer-learned models
outperforming the baseline. Both Euclidean and cosine-based 1-NN
perform well on the many-valued and symbolic view spaces, with no
significant difference between representations or models.
Importantly, while a decision tree classifier was unable to learn
within the many-valued view, the same procedure applied to the
symbolic view produced competitive classification results, a faithful
surrogate for the NN. This demonstrates the value of the symbolic
translation: it transforms the representation into a form that is
amenable to inherently interpretable classifiers.

\begin{table}[t]
  \caption{Five NNs (Tanh) and their (symbolic) conceptual
    views captured by different surrogates (decision tree,
    1-NN). IncV3 could not distinguish \emph{Cherry~1} and
    \emph{Plum}; EffB0 could not distinguish \emph{Apple Red~1} and
    \emph{Apple Pink Lady} in the symbolic view.}
  \centering
\begin{tabular}{|l|c||c|c||c|c||c|c||c|}
    \hline
  Model
  &Model&\multicolumn{2}{c||}{DTree}&\multicolumn{2}{c||}{Euclidean}&\multicolumn{2}{c||}{Cos}&\\\hline
  &ACC&ACC&Fid&ACC&Fid&ACC&Fid&\\
  \hline \hline
  Baseline&0.936&0.017&0.017&0.935&0.989&0.935&0.988&\\
  VGG16&0.988&0.017&0.018&0.988&0.998&0.988&0.997&\\
  ResNet50&0.989&0.018&0.018&0.989&0.998&0.989&0.997&\\
  IncV3&0.983&0.013&0.013&0.983&0.999&0.984&0.999&\\
  EffB0&0.984&0.007&0.007&0.984&0.998&0.983&0.984&\\ \hline \hline
                 \multicolumn{8}{|l|}{Symbolic}&Class Sep\\\hline\hline
  Baseline&0.936&0.857&0.879&0.927&0.964&0.927&0.964&1.0\\
  VGG16&0.988&0.972&0.977&0.988&0.994&0.988&0.994&1.0\\
  ResNet50&0.989&0.952&0.957&0.988&0.996&0.988&0.996&1.0\\
  IncV3&0.983&0.975&0.988&0.984&0.997&0.984&0.997&0.992\\
  EffB0&0.984&0.938&0.934&0.984&0.996&0.984&0.996&0.992\\ \hline
  \end{tabular}
  \label{tab:fruit-fidel-symbolic}
\end{table}

\section{Reasoning with Neuro-Symbolic Ontologies}\label{sec:fca-reasoning}

\paragraph{Symbolic Conceptual Views through FCA.}
Each algebraic relation corresponds to exactly one natural order
structuring of the data by means of the underlying Galois connections.
Through FCA one reveals this order by means of \emph{formal concepts},
i.e., all $(A,B)\in\mathcal{P}(G)\times \mathcal{P}(M)$ such that
$A^{I}=B$ and $B^{I}= A$. The sets $A$ and $B$ are called
\emph{extent} and \emph{intent}, respectively (strongly resembling
maximal bi-cliques). The set of all formal concepts of 
$\context$ is denoted by $\mathfrak{B}(\context)$ and ordered by
inclusion $\subseteq$ on the extent sets. The resulting order structure
$\BV(\context)\coloneqq (\mathfrak{B}(\context),\subseteq)$
constitutes a complete lattice that can be visualized as a \emph{line
  diagram}. Applied to NNs, one may visualize the conceptual hierarchy
learned by the network, and the lattice size serves as an upper bound
for the number of learned concepts.

%

\textbf{Observations.}  The concept lattices for the Baselines,
ResNet50, VGG16, IncV3, and EffB0 models vary in sizes between 126,487
(VGG16) and 134,100 (IncV3) for $|N|=16$, and between 3,498,829
(VGG16) and 3,803,799 (ResNet50) for $|N|=32$. Restricting to positive
attributes (omitting $\bar N$) decreases the sizes by one order
of magnitude: between 5,200 (VGG16) and 6,573 (EffB0) for $|N|=16$,
and 150,884 (EffB0) to 198,152 (IncV3) for $|N|=32$ (see
Appendix,~\cref{tab:Neuronal-lattices}).

\begin{figure*}[b]
  \centering
    \begin{minipage}{0.9\textwidth}
      \begin{tabular}{ccccc}
        \phantom{aaaaaa}Base&
         \phantom{aaaaaaaa}VGG16&
          \phantom{aaaaaaaaa}ResNet&
           \phantom{aaaaaaaa}IncV3&
            \phantom{aaaaaaaa}EffB0
      \end{tabular}
      \includegraphics[width=1\textwidth,trim=4cm 7cm 5cm 9cm,clip]{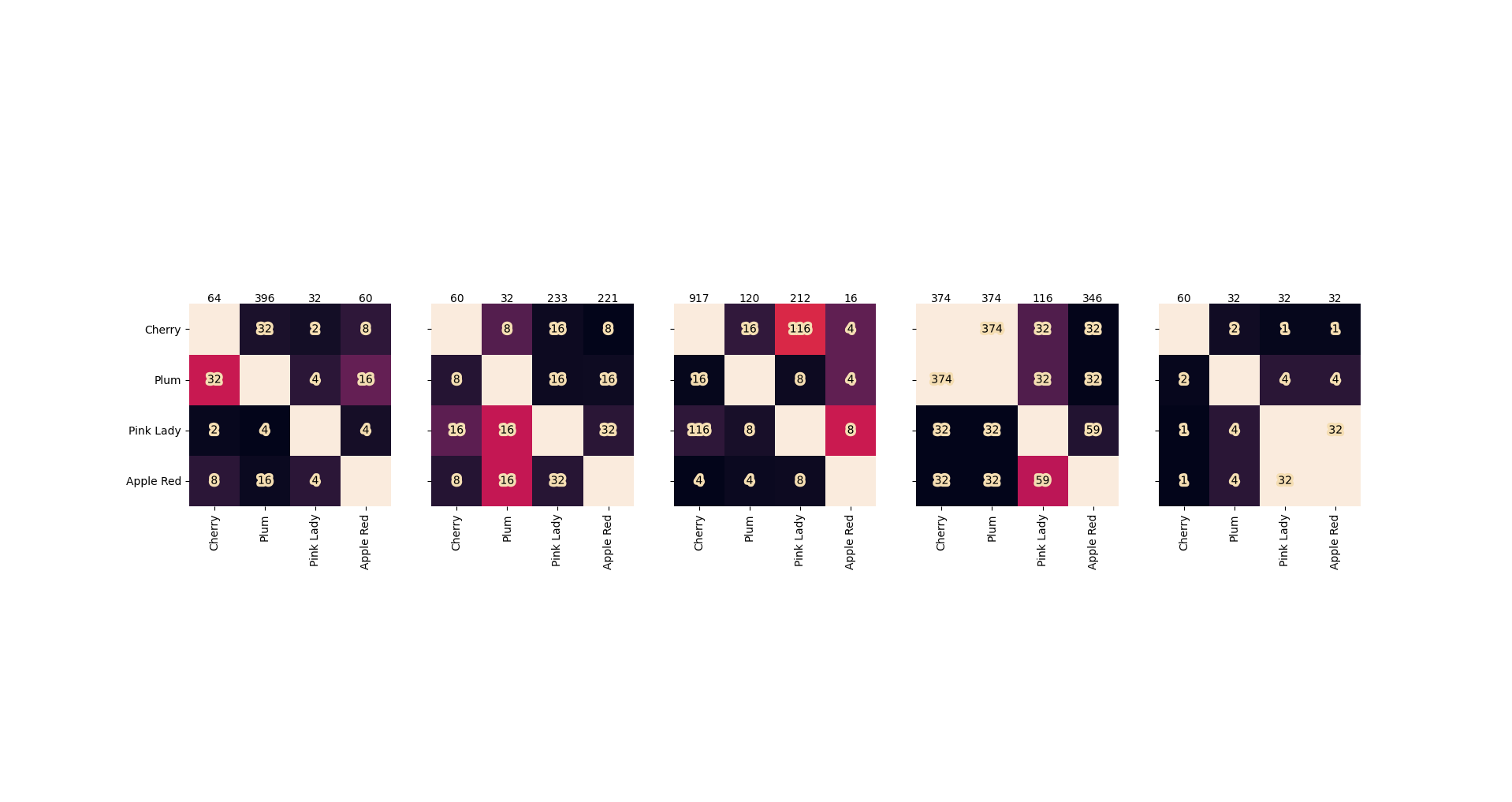}
  \end{minipage}
  \begin{minipage}{0.09\linewidth}
    \scalebox{0.15}{\includegraphics[trim=42cm 0cm 0cm 0cm,clip]{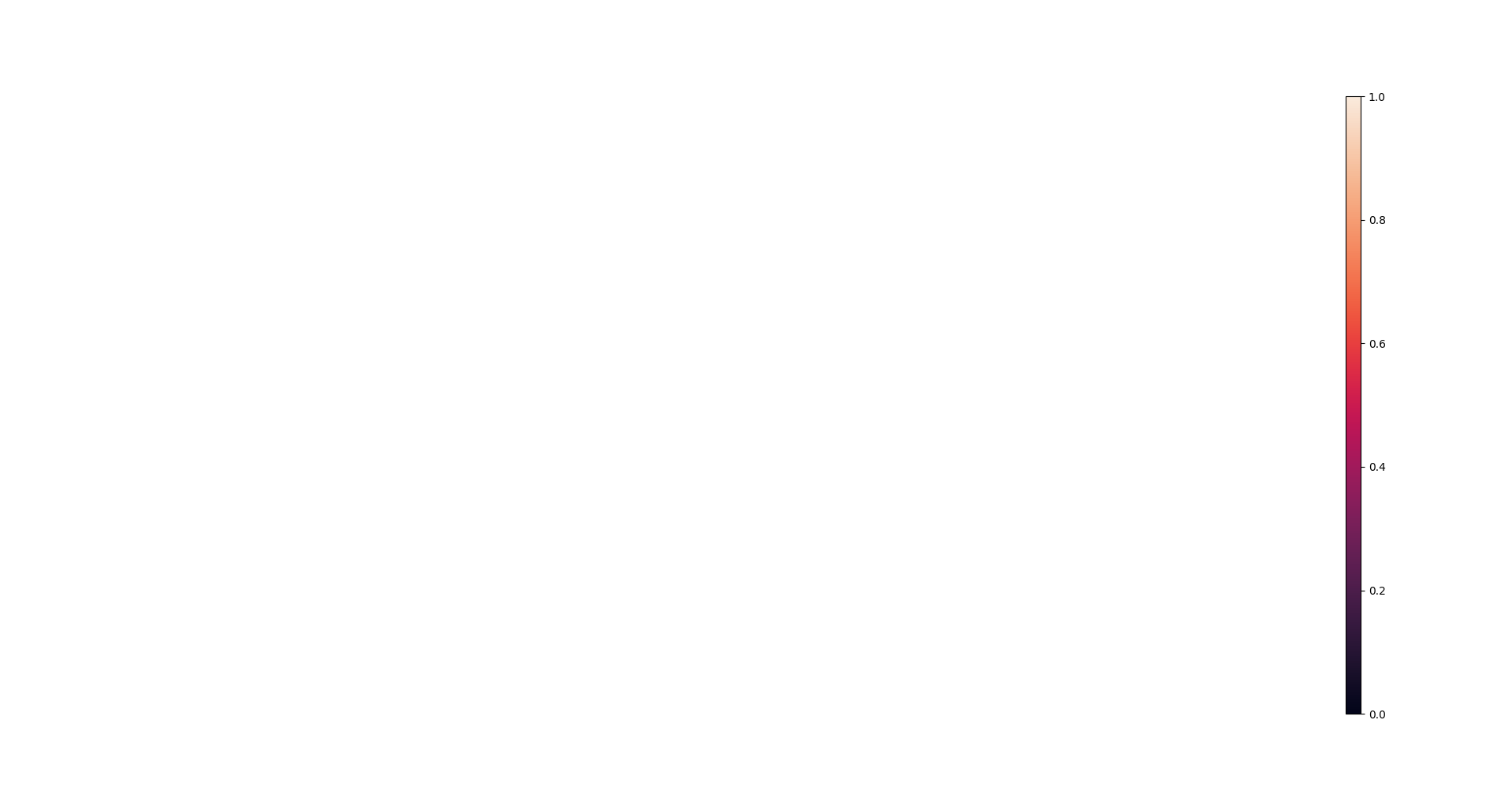}}
  \end{minipage}\\
  \caption{FCA results for \emph{Apple Red}, \emph{Pink Lady},
    \emph{Plum}, and \emph{Cherry} using views with positive
    attributes. Column headers show the number of  concepts. Cell
    values indicate shared concepts between row/column fruits; heat
    indicates the corresponding fraction.}
  \label{fig:neuro-concepts-stats}
\end{figure*}

All lattices are similar in size and too large for direct
visualization. However, the formal concepts encode combinations of
features. The minimum number of encoded features is present in
\emph{meet-irreducible} (MI) elements, whose count is bounded by the
number of attributes and serves as a lower bound on the concepts
captured by the model.
Independent of size, this translation to FCA enables the application
of various knowledge-based methods such as description logics or
subgroup discovery, as investigated in~\cref{sec:abductive}. Within
FCA we can analyze cuts of the lattice, particularly where we suspect
problems in the representation. As we discovered that \emph{Apple Red},
\emph{Pink Lady}, \emph{Plum}, and \emph{Cherry} are
indistinguishable by some symbolic representations
(\cref{tab:fruit-fidel-symbolic}), we ``zoom'' into those.
\cref{fig:neuro-concepts-stats} shows the statistical analysis and
\cref{fig:neuro-concepts} shows the structural analysis.

From \cref{fig:neuro-concepts-stats} we identify formal concept-based
similarities among selected fruits and the number of shared
concepts. For example, Cherry and Plum are indistinguishable in IncV3
while Pink Lady and Apple Red are indistinguishable in EffB0.
Conversely, Cherry and Plum in the ResNet model are quite distinct.
The similarity plots indicate that different models extract different
properties from the data, offering complementary perspectives that
could be combined within the framework.
In the concept lattice (\cref{fig:neuro-concepts}), the reader can
infer hierarchical dependencies between the fruits (indicated by
color), providing structural insight into how the model organizes its
learned representations.

\begin{figure*}[t]
  \centering
  \scalebox{0.35}{\input{pics/concepts/vggplum.tikz}}
  \caption{Order-structured representation of concepts with positive
    attributes containing \emph{Plum}. Concepts containing
    \emph{Apple Pink Lady} and \emph{Apple Red 1} are highlighted in
    orange; \emph{Cherry 1} is highlighted in blue.}
  \label{fig:neuro-concepts}
\end{figure*}

\section{Abductive Learning of Neuro-Symbolic Ontologies}\label{sec:abductive}
Symbolic conceptual views enable the application of various logical
methods to derive human-comprehensible (partial) explanations. We
construct a formal context
$\mathbb{C}=(C,S_{M},I_{\mathbb{C}})$, where $S_{M}=
\{S_{m_1},\dots,S_{m_l}\}$ is a set of human-interpretable features
known about the classes $C$, i.e., \emph{background knowledge}.

\begin{definition}[Symbolic Interpretation]
  Given the symbolic conceptual view
  $\mathcal{V}_{\mathbb{D}}=(\mathbb{O}_{\mathbb{D},}\mathbb{W}_{\mathbb{D}})$
  of a NN, background knowledge $\mathbb{C}$, and a similarity
  relation $\sim$ on the classes $\mathcal{P}(C)$. Then the formal context
  $\mathbb{S}=(N,S_M,R)$ with $(n,S_{m})\in R \longeq
  \{n\}^{I_{\mathbb{W}}}\sim \{S_{m}\}^{I_{\mathbb{C}}}$ is the
\emph{symbolic interpretation} of the NN wrt $\mathbb{C}$
and $\sim$.
\end{definition}

We require $\sim$ to be reflexive and symmetric but not necessarily
transitive. Symbolically interpreting a NN is to deduce
$\sim$ using background knowledge. Given a symbolic interpretation, we
can express neurons in terms of human-interpretable features $S_M$ by
applying the incidence relation in $\mathbb{S}$, i.e., for all
$n\in N$ one can compute $\{n\}^{R}$. Furthermore, if $\mathbb{S}$ is
equipped with propositional logic
$\mathcal{L}(S_M,\{\vee,\wedge,\neg\})$, then FCA~\cite{smeasure}
provides the means for expressing neurons in terms of propositional
statements.

\begin{figure}[t]
  \centering
\hspace{-0.5cm}
  \includegraphics[width=0.38\textwidth]{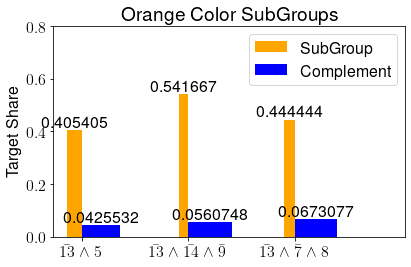}
  \includegraphics[width=0.38\textwidth]{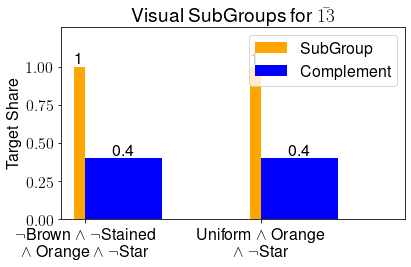}\\
\hspace{-0.5cm}
  \includegraphics[width=0.38\textwidth]{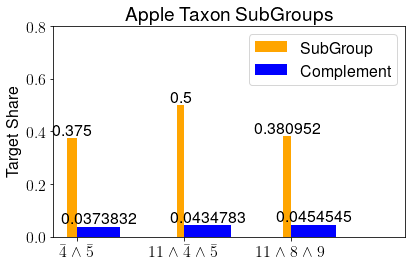}
  \includegraphics[width=0.38\textwidth]{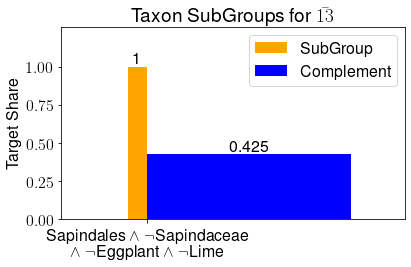}
  \caption{Exemplary results of the subgroup detection. Width indicates
    subgroup size.}
  \label{fig:interpretation-fruit360}
\end{figure}

\textbf{Evaluating the Symbolic View Interpretation.}
We demonstrate how symbolic interpretations can be used to relate
neurons to human-comprehensible features $S_{M}$ and vice versa,
analyzing the symbolic view of the Fruits-360 data set. For attributes
$S_{M}$ we use: (1) visual features such as shapes and colors, and (2)
the \emph{Scientific classification} taxonomy from
Wikipedia\footnote{See e.g.\
  \url{https://en.wikipedia.org/wiki/Apple}} for each
fruit/vegetable, combining German and English articles. We infer the
similarity relation using \emph{subgroup
  discovery}~\cite{subgroup}, as implemented in \texttt{pysubgroup}.


We depict four example results
in~\cref{fig:interpretation-fruit360}. On the left we show
subgroups in terms of neurons; on the right, in terms of interpretable
features. On the abscissa we find propositional statements combining
the respective features.
From the high share (ordinate) of the respective subgroups we can
infer that the propositional statements describe the taxons from
adequately up to very well. In the latter case (right), the subgroups
are pure yet not complete. Two concrete examples: (1) fruits
that are not brown, not stained, not orange, and not star-shaped will
activate neuron $\bar{13}$ from $\bar N$; (2) if neurons $\bar{13}$,
$\bar{14}$, and $\bar{9}$ are activated, this implies the fruit is
orange with confidence $\approx 0.54$. Using this method one can infer
the similarity relation $\sim$ and provide a structured explanation
framework.

\section{Discussion and Limitations}\label{sec:discussion}

\textbf{Summary of Contributions.}
We introduced \emph{conceptual views} as a principled framework for
the global analysis of neural networks, grounded in the algebraic
foundations of Formal Concept Analysis. Our framework is distinct from
prior work in three key aspects: (1) it does not rely on additional
opaque methods such as autoencoders for explanation; (2) it is global
by design, analyzing the entire model rather than individual
predictions; and (3) it does not require pre-defined concepts or their
associated input representations. We accomplish this by jointly
analyzing both the weights of the output layer and the activations of
the last hidden layer through symbolic conceptual views.

\textbf{Experimental Validity.}
Our experimental study encompasses twenty-four diverse ImageNet models
and controlled experiments on the Fruits-360 data set.  We
deliberately structured the experiments to validate each stage of the
framework independently: the many-valued view's fidelity (up to
0.999), the symbolic view's capacity for class separation (uniformly
1.0), and the interpretability of derived rules. We emphasize that our
experiments are designed as a proof of concept for the framework rather
than as a competitive benchmark; the primary contribution is the
formal methodology, not the specific numerical results.

\textbf{Limitations.}
We identify several limitations that should guide future work:
\begin{inparadesc}
\item[Restriction to non-recursive architectures.] Our
  current formalization assumes feed-forward networks with a
  well-defined last hidden layer. Adapting the framework to recurrent
  or attention-based architectures (e.g., Transformers) would require
  a substantial extension of the view definition, potentially
  involving time-unrolled or attention-aggregated representations.
\item[Multiple outputs requirement.] The class view requires
  multiple output neurons. While common splitting techniques can
  address single-output scenarios, the effectiveness of conceptual
  views in regression settings remains unexplored.
\item[Dependence on background knowledge.] Human-comprehensible
  explanations require domain-specific background knowledge. The
  quality of the abductive interpretations is bounded by the richness
  and accuracy of this knowledge. Automating background knowledge
  acquisition, e.g., through knowledge graphs or large language
  models, is a promising direction.
\item[Threshold sensitivity.] Our parameter study shows that
  the quality of the symbolic view depends heavily on the activation
  function, with Tanh strongly preferred over ReLU for dichotomic
  scaling. More sophisticated scaling strategies (such as multi-level
  or data-driven thresholding) could improve results for
  architectures that use non-symmetric activation functions.
\item[Scalability] The lattices grow exponentially
  with the number of attributes. This does not prevent
  computational analysis (e.g., implications can be computed without
  materializing the full lattice), but it limits direct human inspection.
\end{inparadesc}

\textbf{Scope of Experiments.}  Our experimental study focuses on
image classification tasks using established benchmark data
sets. Extending the evaluation to other modalities (text, tabular,
time series) and to more recent architectures (Vision Transformers,
diffusion models) would strengthen the generality claims of the
framework. Due to the page limit of these proceedings, we defer such
extensions to future work. We note, however, that the formal
definitions are architecture-agnostic: any model with a
distinguishable last hidden layer and a subsequent classification
layer can be analyzed through conceptual views.

\textbf{Future Directions.}
The presented link between NN models and FCA through symbolic
conceptual views enables both improved explainability and potential
enhancements of surrogate learning procedures. Particularly promising
directions include: integrating the framework with concept bottleneck
models~\cite{cbm-koh2020} to combine architectural and post-hoc
interpretability; extending the symbolic view to intermediate layers
for a multi-resolution analysis; and exploring whether feedback from
the symbolic analysis can guide model refinement, e.g., by identifying
neurons that encode redundant or biased concepts.
\bibliographystyle{splncs04}
\bibliography{paper}

\newpage
\appendix
\section{Appendix}
In this appendix we provide additional figures and tables for our
experimental studies.

\begin{table}
  \caption{Fidelity between twenty-four
    neural networks and their object/class view using 1-NN.}
  \centering
    \begin{tabular}{|l||c|c|}
      \hline
      Model&Euclidean&Cosine\\ \hline \hline
      VGG16&0.945&0.841\\
      VGG19&0.942&0.842\\
      IncV3&0.990&0.753\\
      DenseNet121&0.978&0.737\\
      DenseNet169&0.989&0.843\\
      DenseNet201&0.972&0.728\\
      MobilNetV1&0.575&0.449\\
      MobilNetV2&0.947&0.925\\
      NasNetMobile&0.935&0.808\\
      NasNetLarge&0.880&0.831\\
      ResNet50&0.954&0.800\\
      ResNet50v2&0.995&0.734\\
      \hline
    \end{tabular}
    \begin{tabular}{|l||c|c|}
      \hline
      Model&Euclidean&Cosine\\ \hline \hline
      ResNet101V2&0.995&0.466\\
      ResNet152V2&0.999&0.314\\
      IncResNetV2&0.999&0.983\\
      XCeption&0.977&0.792\\
      EffB0&0.944&0.933\\
      EffB1&0.960&0.946\\
      EffB2&0.969&0.957\\
      EffB3&0.974&0.961\\
      EffB4&0.981&0.972\\
      EffB5&0.979&0.972\\
      EffB6&0.982&0.976\\
      EffB7&0.985&0.979\\
      \hline
    \end{tabular}
  \label{tab:imagenet-fidel-long}
\end{table}

\begin{table}
  \caption{Average weights $w_{i,j}$, object values $n_i(g)$, number
    of neurons $|N|$, and activation function $f$ of the last hidden
    layer for all TensorFlow ImageNet models.}
  \centering
  \begin{tabular}{|l||r|r||r||c|c|}
    \hline
    Model&\multicolumn{1}{c|}{$\mathbb{W}$ - values}&\multicolumn{1}{c||}{Bias}&\multicolumn{1}{c||}{$\mathbb{O}$ - values}&$|N|$&$f$\\ \hline \hline
    VGG16&-5.359e-07 $\pm$ 0.008&1.404e-06 $\pm$ 0.191&0.679 $\pm$ 1.514&4096&ReLU\\
    VGG19&-6.707e-07 $\pm$ 0.008&-1.287e-05 $\pm$ 0.192&0.613 $\pm$ 1.402&4096&ReLU\\
    IncV3&-3.808e-05 $\pm$ 0.034&-0.0099 $\pm$ 0.308&6.025 $\pm$ 15.13&2048&ReLU\\
    DenseNet121&2.139e-08 $\pm$ 0.049&-1.014e-07 $\pm$ 0.012&1.731 $\pm$ 4.603&1024&ReLU\\
    DenseNet169&1.456e-08 $\pm$ 0.039&-1.038e-07 $\pm$ 0.012&1.675 $\pm$ 5.529&1664&ReLU\\
    DenseNet201&1.019e-08 $\pm$ 0.036&-1.178e-07 $\pm$ 0.011&1.146 $\pm$ 4.167&1920&ReLU\\
    MobilNetV1&-0.0001 $\pm$ 0.081&-0.005 $\pm$ 0.744&0.435 $\pm$ 0.838&1024&ReLU\\
    MobilNetV2&-3.138e-05 $\pm$ 0.041&0.0002 $\pm$ 0.319&0.358 $\pm$ 0.747&1280&ReLU\\
    NasNetLarge&-2.080e-07 $\pm$ 0.026&4.424e-05 $\pm$ 0.040&0.198 $\pm$ 0.533&4032&ReLU\\
    NasNetMobile&-3.336e-07 $\pm$ 0.039&0.0001 $\pm$ 0.066&0.382 $\pm$ 4.389&1056&ReLU\\
    ResNet50&3.774e-07 $\pm$ 0.033&-4.881e-08 $\pm$ 0.009&0.546 $\pm$ 0.871&2048&ReLU\\
    ResNet101V2&6.668e-06 $\pm$ 0.027&0.0016 $\pm$ 0.292&39.97 $\pm$ 167.8&2048&ReLU\\
    ResNet152V2&1.038e-05 $\pm$ 0.026&0.0016 $\pm$ 0.287&94.08 $\pm$ 187.4&2048&ReLU\\
    ResNet50V2&8.014e-07 $\pm$ 0.028&0.0011 $\pm$ 0.292&19.91 $\pm$ 74.65&2048&ReLU\\
    IncResNetV2&-3.060e-05 $\pm$ 0.037&-0.0012 $\pm$ 0.230&106.8 $\pm$ 124.9&1536&ReLU\\
    XCeption&-3.246e-06 $\pm$ 0.055&0.0008 $\pm$ 0.281&2.974 $\pm$ 13.41&2048&ReLU\\
    EffB0&-7.495e-05 $\pm$ 0.068&-5.143e-05 $\pm$ 0.058&0.065 $\pm$ 0.321&1280&Swish\\
    EffB1&-5.647e-05 $\pm$ 0.063&-4.343e-05 $\pm$ 0.045&0.056 $\pm$ 0.313&1280&Swish\\
    EffB2&-7.152e-05 $\pm$ 0.059&-4.153e-05 $\pm$ 0.054&0.019 $\pm$ 0.260&1408&Swish\\
    EffB3&-6.323e-05 $\pm$ 0.054&-3.547e-05 $\pm$ 0.046&0.010 $\pm$ 0.252&1536&Swish\\
    EffB4&-3.106e-05 $\pm$ 0.050&-3.138e-05 $\pm$ 0.057&-0.039 $\pm$ 0.194&1792&Swish\\
    EffB5&-2.043e-05 $\pm$ 0.049&-2.738e-05 $\pm$ 0.055&-0.036 $\pm$ 0.170&2048&Swish\\
    EffB6&-8.656e-06 $\pm$ 0.046&-2.691e-05 $\pm$ 0.071&-0.043 $\pm$ 0.135&2304&Swish\\
    EffB7&-9.562e-06 $\pm$ 0.041&-2.441e-05 $\pm$ 0.060&-0.041 $\pm$ 0.136&2560&Swish\\
\hline
  \end{tabular}
  \label{tab:weights-long}
\end{table}

\begin{figure}[b]
  \centering
  \begin{minipage}{0.49\linewidth}
    \includegraphics[width=1\textwidth]{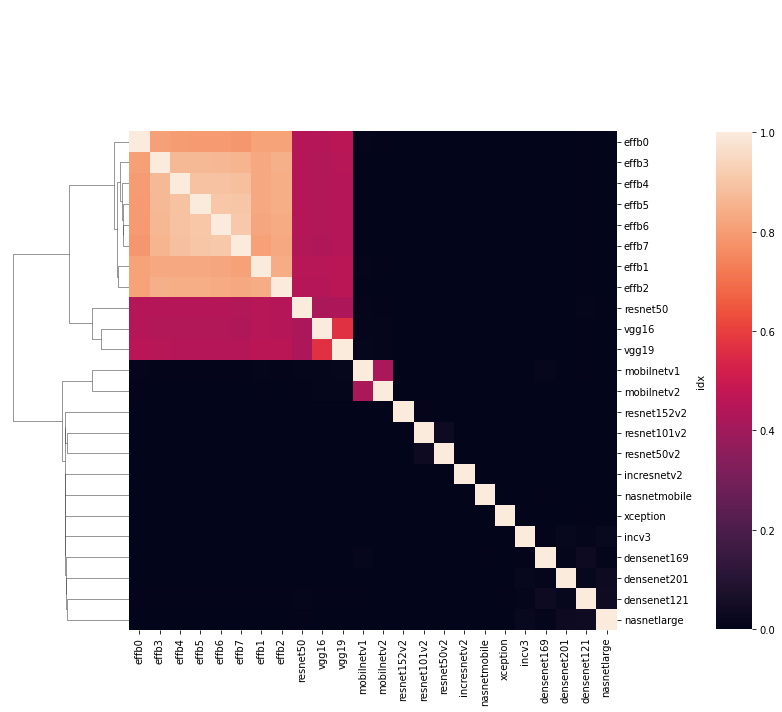}
  \end{minipage}
  \begin{minipage}{0.5\linewidth}
    \includegraphics[width=\textwidth]{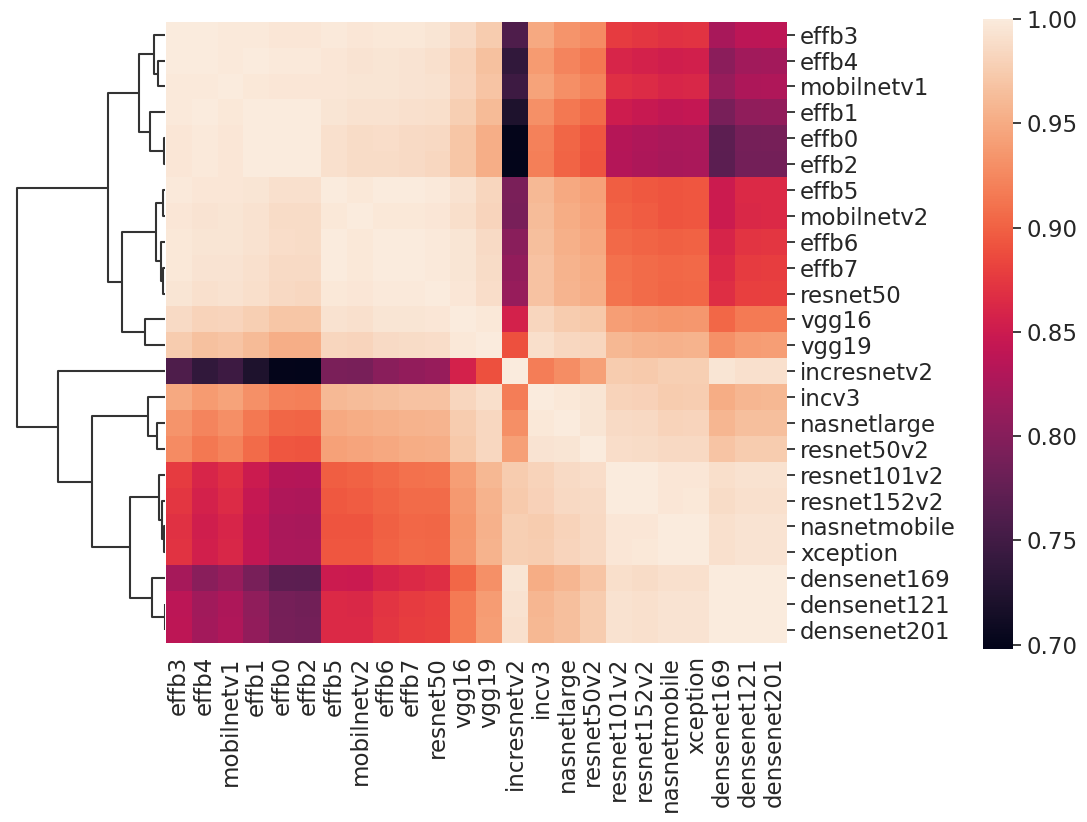}\\
  \includegraphics[width=\textwidth]{pics/imagenet_euclid_gromov_dendrogram.png}
  \end{minipage}
  \caption{Similarity of twenty-four neural networks trained on
    ImageNet. Left: pairwise fidelity baseline. Right:
    Gromov-Wasserstein distance on the object (top) and class (bottom)
    view.}
  \label{fig:imagenet-sim-long}
\end{figure}

\begin{figure}
 \phantom{,}  \hspace{0.5\textwidth}\textbf{Swish}\\
  \begin{tabular}{lm{0.2\textwidth}}
    $\mathbb{O}$& \includegraphics[width=0.9\textwidth]{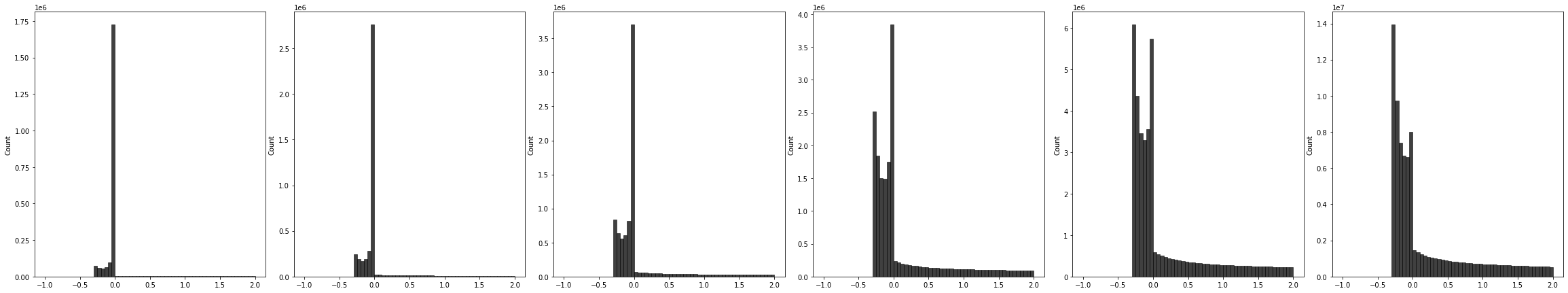}\\
    $\mathbb{W}$& \includegraphics[width=0.9\textwidth]{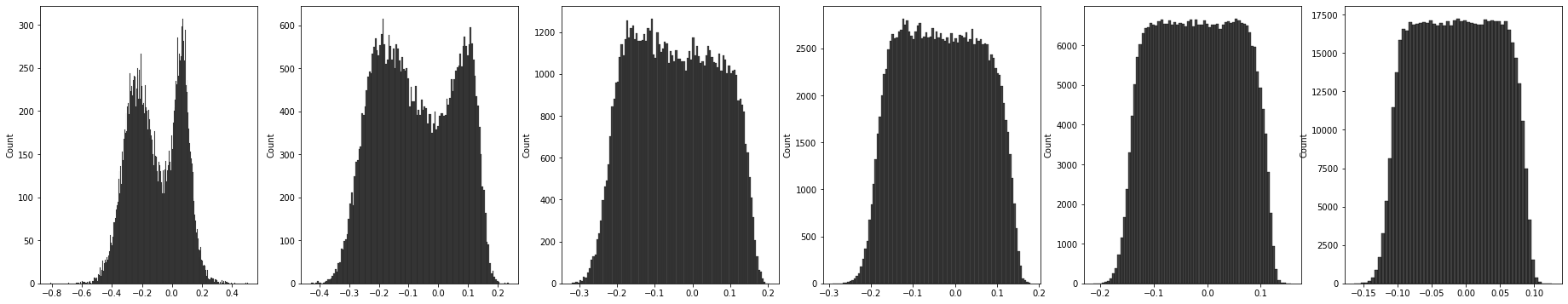}
  \end{tabular}\\
 \phantom{,} \hspace{0.5\textwidth} \textbf{ReLU}\\
  \begin{tabular}{lm{0.2\textwidth}}
    $\mathbb{O}$& \includegraphics[width=0.9\textwidth]{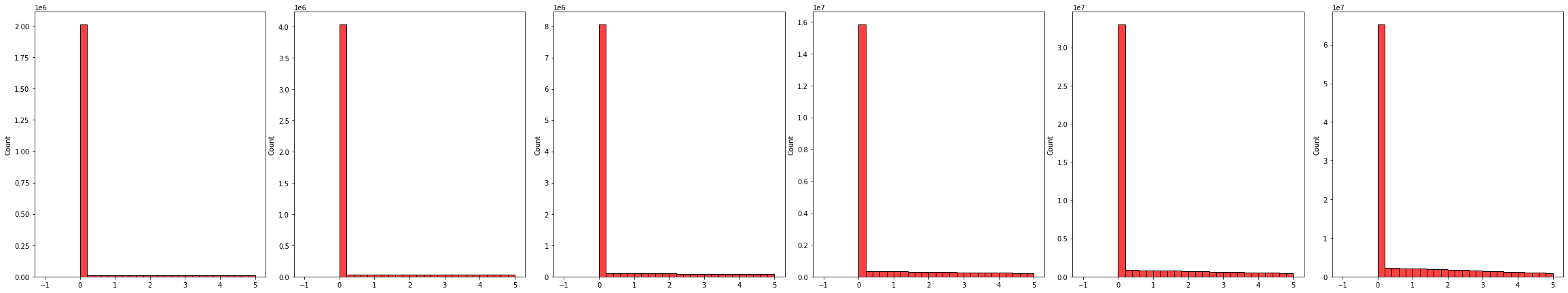}\\
    $\mathbb{W}$& \includegraphics[width=0.9\textwidth]{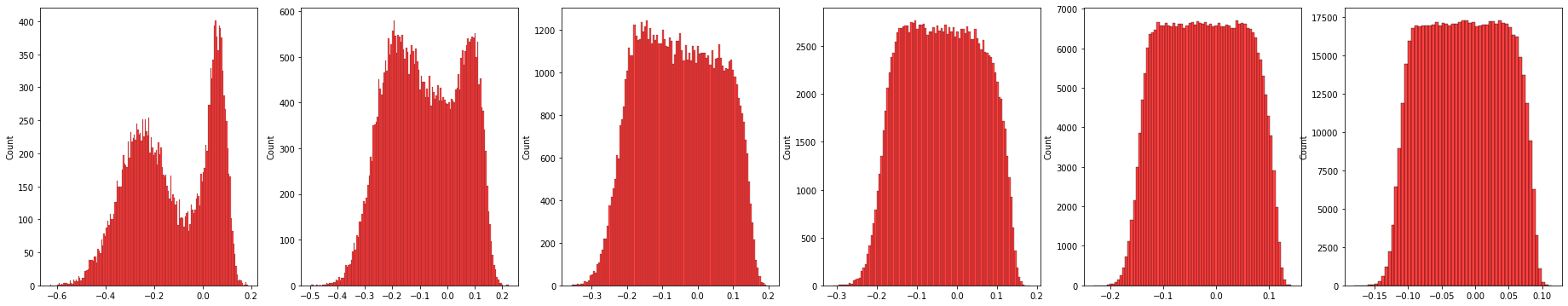}
  \end{tabular}\\
 \phantom{,}  \hspace{0.5\textwidth}\textbf{Linear}\\
  \begin{tabular}{lm{0.2\textwidth}}
    $\mathbb{O}$& \includegraphics[width=0.9\textwidth]{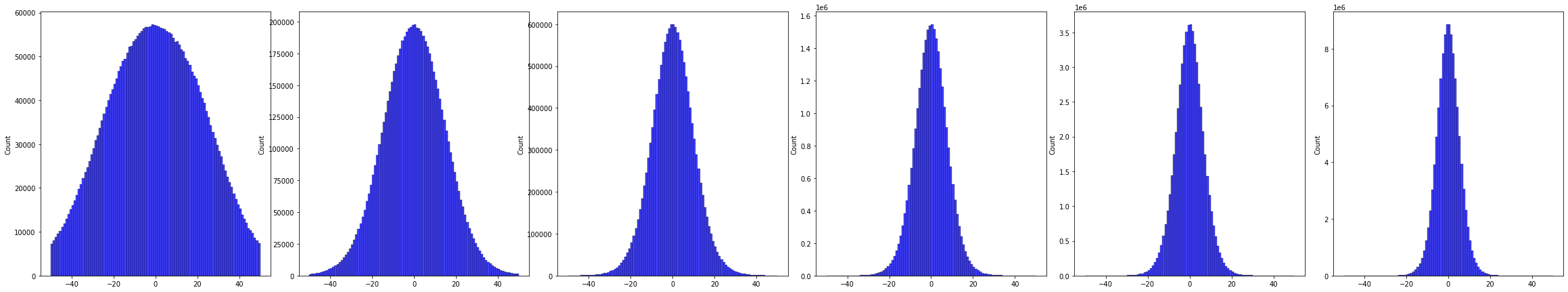}\\
    $\mathbb{W}$& \includegraphics[width=0.9\textwidth]{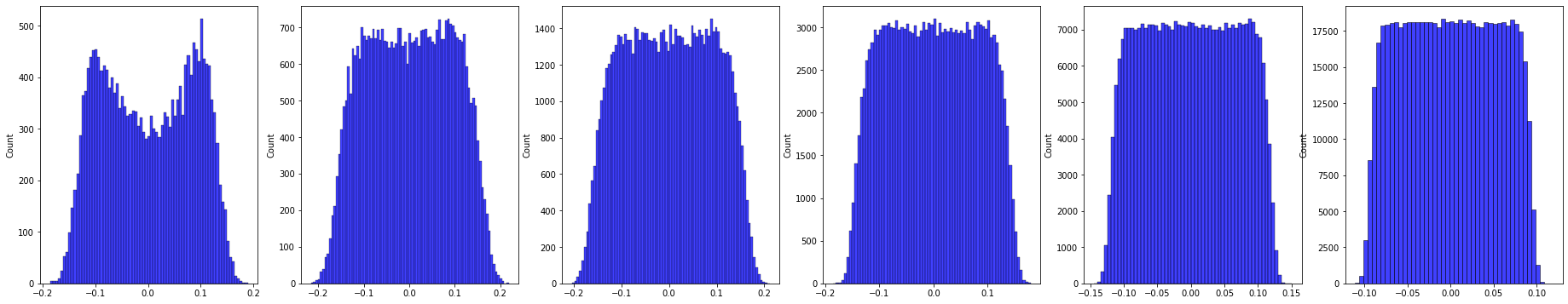}
  \end{tabular}\\
 \phantom{,}  \hspace{0.5\textwidth} \textbf{TanH}\\
  \begin{tabular}{lm{0.2\textwidth}}
    $\mathbb{O}$& \includegraphics[width=0.9\textwidth]{pics/activation/tanh-all-O.png}\\
    $\mathbb{W}$& \includegraphics[width=0.9\textwidth]{pics/activation/tanh-all.png}
  \end{tabular}
  \caption{Value distributions for the object ($\mathbb{O}$) and class
    ($\mathbb{W}$) view for ten runs using Fruits-360 with Swish,
    ReLU, Linear, and Tanh activation. Last hidden layer from $2^{4}$
    (first column) to $2^{9}$ (last column).}
  \label{fig:weights-long}
\end{figure}

\begin{table}
  \caption{Full parameter study: influence of activation function and
    neuron count on the quality of the (symbolic) object/class view.
    V-Fid and SV-Fid denote fidelity of 1-NN to the original model.}
\centering
\begin{tabular}{|l||c|c|c|c|c|c|}
    \hline
    &$2^{4}$& $2^{5}$&$2^{6}$&$2^{7}$&$2^{8}$&$2^{9}$\\ \hline\hline
    \multicolumn{7}{|l|}{\textbf{Swish}}\\ \hline
       $\delta_{\mathbb{O}}=0$ Split                             &57.1/62.9  &52.8/47.2  &49.2/50.8  &44.5/55.5  &45.6/54.4  &45.0/55.0\\
       $\delta_{\mathbb{W}}=0$ Split                             &65.4/44.6  &65.0/35.0  &62.0/38.0  &60.1/39.9  &57.2/43.8  &56.5/43.5\\
                       Model Acc                                &93.5$\pm$ 0.8&94.5$\pm$ 0.5&95.3$\pm$ 0.3&95.1$\pm$ 0.3&95.4$\pm$ 0.4&95.1$\pm$ 0.5\\
    V-Fid                                                       &99.5$\pm$ 0.4&99.9$\pm$ 0.0&99.9$\pm$ 0.0&99.9$\pm$ 0.0&99.9$\pm$ 0.0&99.9$\pm$ 0.0\\
    SV-Fid                                                      &77.1$\pm$ 9.2&88.8$\pm$ 1.3&89.6$\pm$ 1.6&88.8$\pm$ 1.4&88.4$\pm$ 0.9&86.7$\pm$ 1.6\\ \hline
    \multicolumn{7}{|l|}{\textbf{ReLU}}\\\hline
       $\delta_{\mathbb{O}}=0$ Split                             &55.1/44.9  &55.1/44.9  &54.7/45.3  &53.3/46.7  &55.3/44.7  &54.1/45.9\\
       $\delta_{\mathbb{W}}=0$ Split                             &66.3/33.7  &66.7/33.2  &64.0/36.0  &61.6/38.4  &58.9/41.1  &58.2/41.8\\
                       Model Acc                                &93.7$\pm$ 0.4&94.5$\pm$ 0.5&94.9$\pm$ 0.5&94.9$\pm$ 0.5&95.0$\pm$ 0.4&94.8$\pm$ 0.6\\
    V-Fid                                                       &99.7$\pm$ 0.0&99.8$\pm$ 0.0&99.9$\pm$ 0.0&99.9$\pm$ 0.0&99.9$\pm$ 0.0&99.9$\pm$ 0.0\\
    SV-Fid                                                      &79.9$\pm$ 3.7&89.0$\pm$ 1.2&90.0$\pm$ 1.2&89.5$\pm$ 1.1&89.0$\pm$ 1.5&88.2$\pm$ 1.5\\ \hline
\end{tabular}
\begin{tabular}{|l||c|c|c|c|c|c|}
    \hline
    &$2^{4}$& $2^{5}$&$2^{6}$&$2^{7}$&$2^{8}$&$2^{9}$\\ \hline\hline
    \multicolumn{7}{|l|}{\textbf{Linear}}\\ \hline
       $\delta_{\mathbb{O}}=0$ Split                             &49.8/50.2  &49.6/50.4  &49.5/50.5  &49.9/50.1  &49.9/50.1  &49.9/50.1\\
       $\delta_{\mathbb{W}}=0$ Split                             &49.7/50.3  &49.3/50.7  &49.7/50.3  &50.0/50.0  &50.0/50.0  &50.0/50.0\\
                       Model Acc                                &85.3$\pm$ 0.6&88.8$\pm$ 0.7&89.9$\pm$ 1.0&92.0$\pm$ 0.5&91.8$\pm$ 0.9&91.5$\pm$ 0.9\\
    V-Fid                                                       &99.9$\pm$ 0.0&99.9$\pm$ 0.0&99.9$\pm$ 0.0&99.9$\pm$ 0.0&99.9$\pm$ 0.0&99.9$\pm$ 0.0\\
    SV-Fid                                                      &55.5$\pm$ 1.9&64.7$\pm$ 1.4&68.3$\pm$ 2.6&74.8$\pm$ 1.7&78.2$\pm$ 2.1&81.5$\pm$ 1.1\\ \hline
    \multicolumn{7}{|l|}{\textbf{Tanh}}\\\hline
       $\delta_{\mathbb{O}}=0$ Split                             &49.7/50.3  &49.7/50.3  &49.8/50.2  &49.9/50.1  &50.0/50/0  &49.9/50.1\\
       $\delta_{\mathbb{W}}=0$ Split                             &49.9/50.1  &49.8/50.2  &49.8/50.2  &50.0/50.0  &49.9/50.1  &50.0/50.0\\
                       Model Acc                                &90.5$\pm$ 0.8&94.3$\pm$ 0.5&94.7$\pm$ 0.5&94.9$\pm$ 0.4&95.0$\pm$ 0.4&94.8$\pm$ 0.3\\
    V-Fid                                                       &98.3$\pm$ 0.5&99.5$\pm$ 0.1&99.7$\pm$ 0.0&99.7$\pm$ 0.0&99.8$\pm$ 0.0&99.8$\pm$ 0.0\\
    SV-Fid                                                      &94.3$\pm$ 1.4&97.4$\pm$ 0.4&97.7$\pm$ 0.4&97.6$\pm$ 0.1&97.8$\pm$ 0.2&97.6$\pm$ 0.2\\ \hline

  \end{tabular}
  \label{tab:distro-stats-long}
\end{table}

\begin{figure}
  \centering
  \begin{tabular}{cc}
    \textbf{EffB7}&\textbf{VGG16}\\
    \includegraphics[width=0.4\textwidth]{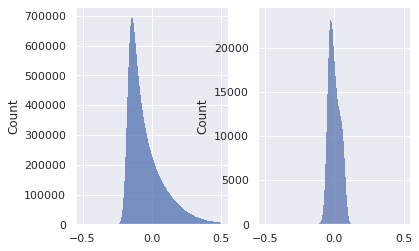}&\includegraphics[width=0.4\textwidth]{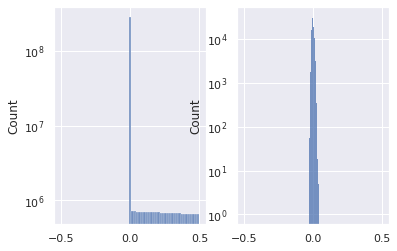}\\
    \textbf{ResNet50}&\textbf{ResNet150V2}\\
    \includegraphics[width=0.4\textwidth]{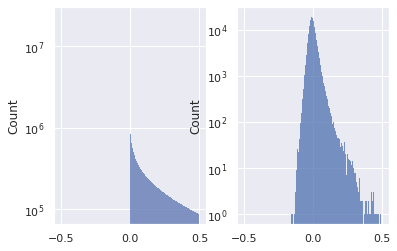}&\includegraphics[width=0.4\textwidth]{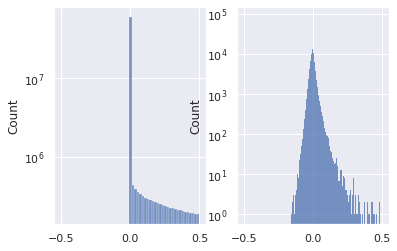}\\
  \end{tabular}
  \caption{Value distributions for the object/class view for EffB7, VGG16, ResNet50, and ResNet150V2.}
  \label{fig:imagenet-distros-long}
\end{figure}

\begin{table}
  \caption{Fidelity between twenty-four neural networks and their
    symbolic object/class view using nearest neighbor and cosine
    similarity for classification.}
  \centering
    \begin{tabular}{|c||c|c|c|c|}
      \hline
      &1NN&Cos&Class Sep&Activation\\ \hline \hline
      VGG16&0.552&0.552&1.0&ReLU\\
      VGG19&0.5672&0.5672&1.0&ReLU\\
      IncV3&0.000&0.000&1.0&ReLU\\
      DenseNet121&0.000&0.000&1.0&ReLU\\
      DenseNet169&0.001&0.001&1.0&ReLU\\
      DenseNet201&0.000&0.000&1.0&ReLU\\
      MobilNetV1&0.036&0.036&1.0&ReLU\\
      MobilNetV2&0.305&0.305&1.0&ReLU\\
      NasNetMobile&0.004&0.004&1.0&ReLU\\
      NasNetLarge&0.009&0.009&1.0&ReLU\\
      ResNet50&0.007&0.000&1.0&ReLU\\
      ResNet50V2&0.001&0.003&1.0&ReLU\\
      \hline
    \end{tabular}
    \begin{tabular}{|c||c|c|c|c|}
      \hline
      &1NN&Cos&Class Sep&Activation\\ \hline \hline
      ResNet101V2&0.012&0.012&1.0&ReLU\\
      ResNet152V2&0.000&0.000&1.0&ReLU\\
      IncResNetV2&0.000&0.000&1.0&ReLU\\
      XCeption&0.220&0.220&1.0&ReLU\\
      EffB0&0.758&0.758&1.0&Swish\\
      EffB1&0.813&0.813&1.0&Swish\\
      EffB2&0.869&0.869&1.0&Swish\\
      EffB3&0.898&0.898&1.0&Swish\\
      EffB4&0.929&0.929&1.0&Swish\\
      EffB5&0.935&0.935&1.0&Swish\\
      EffB6&0.951&0.951&1.0&Swish\\
      EffB7&0.957&0.957&1.0&Swish\\
      \hline
    \end{tabular}
  \label{tab:imagenet-fidel-symbolic-long}
\end{table}

\begin{table}\label{tab:Neuronal-lattices}
  \small\sf\centering
  \centering
  \caption{Concept lattice sizes for the symbolic views in
    \cref{tab:fruit-fidel-symbolic}, for all and only positive
    attributes.}
  \begin{tabular}{|l||r|r||r|r|}
    \hline
    Model&\multicolumn{2}{c||}{$|N|=16$}& \multicolumn{2}{c|}{$|N|=32$}\\
         &All&Pos&All&Pos \\
    \hline
    Base&130969&6517&3192044&155416\\
    ResNet50&133130&5872&3803799&165009\\
    VGG16&126487&5200&3498829&193516\\
    IncV3&134100&5670&3782226&198152\\
    EffB0&132403&6573&3767964&150884\\
    \hline
  \end{tabular}
\end{table}

\end{document}

%% file: math_commands.tex
\usepackage{amsmath,amsfonts,bm}

\def\eqref#1{equation~\ref{#1}}

\def\1{\bm{1}}

\DeclareMathAlphabet{\mathsfit}{\encodingdefault}{\sfdefault}{m}{sl}
\SetMathAlphabet{\mathsfit}{bold}{\encodingdefault}{\sfdefault}{bx}{n}



%% file: pics/pipline.tikz
    \begin{tikzpicture}
      \node at (0,2.4) {Neural Network}; 
      \node[rectangle, draw=black] at (0,0) {\input{pics/fnn.tikz}}; 
      \draw[->,thick] (2.35,0) -- (2.85,0);
      \node[rectangle, draw=black,minimum width=4.1cm,minimum height=3.5cm,] at (5,0){};
      \node[] at (5,-0.7) {
\scalebox{0.55}{
\begin{tabular}{|c||c|c|c|c|}
        \hline
        \color{red} $\mathbb{O}(\text{NN})$ &$n_1$&$n_2$& \dots &$n_h$\\ \hline \hline
        $o_1$&\cellcolor{black!52!white}$n_1(o_1)$&\cellcolor{black!35!white}$n_2(o_1)$&&\cellcolor{black!56!white}$n_k(o_1)$\\ \hline
        $o_2$&\cellcolor{black!45!white}$n_1(o_2)$&\cellcolor{black!10!white}$n_2(o_2)$&&\cellcolor{black!34!white}$n_k(o_2)$\\ \hline
        \dots&&&\phantom{$n_k(o_t)$}&\\ \hline
        $o_t$&\cellcolor{black!9!white}$n_1(o_t)$&\cellcolor{black!20!white}$n_2(o_t)$&&\cellcolor{black!69!white}$n_k(o_t)$\\ \hline
\end{tabular}}};
\node[] at (5,0.7) {
\scalebox{0.55}{
\begin{tabular}{|c||c|c|c|c|}
        \hline
  \color{blue} $\mathbb{W}(\text{NN})$     &$n_1$&$n_2$&\dots&$n_h$\\ \hline \hline
        $c_1$&\cellcolor{black!21!white}$w_{1,1}$&\cellcolor{black!76!white}$w_{1,2}$&&\cellcolor{black!85!white}$w_{1,h}$\\ \hline
        $c_2$&\cellcolor{black!65!white}$w_{2,1}$&\cellcolor{black!10!white}$w_{2,2}$&&\cellcolor{black!44!white}$w_{2,h}$\\ \hline
        \dots &\phantom{$n_k(o_t)$}&\phantom{$n_k(o_t)$}&\phantom{$n_k(o_t)$}&\phantom{$n_k(o_t)$}\\ \hline
        $c_k$&\cellcolor{black!9!white}$w_{k,1}$&\cellcolor{black!20!white}$w_{k,2}$&&\cellcolor{black!69!white}$w_{k,h}$\\ \hline
\end{tabular}}
};
       \node[thick,rectangle,draw=red,minimum width=3cm,minimum height=2.9cm] at (-0.6,-0.25){};
       \node[thick,rectangle,draw=blue,minimum width=2.cm,minimum height=2.6cm] at (1,-0.3){};

      \node at (5,2.4) {MV Conc. View};

 \draw[->,thick] (7.2,0) -- (8,0);
 \node[] at (7.5,0.2) {\footnotesize $\delta_{\mathbb{W}}$};
  \node[] at (7.5,-0.2) {\footnotesize $\delta_{\mathbb{O}}$};
\begin{scope}[xshift=9.5cm]
 \node at (0,2.4) {Symbolic Conc. View}; 
\node[] at (0,1.48) {
\scalebox{0.55}{
\begin{tabular}{|c||c|c|c|c|}
        \hline
  \color{blue} $\mathbb{W}_{\mathbb{D}}(\text{NN})$&$n_1$&$n_2$&\dots&$n_h$\\ \hline \hline
        $c_1$&&$\times$&&$\times$\\ \hline
        $c_2$&$\times$&&&$\times$\\ \hline
        \dots&&\phantom{$\times$}&&\\ \hline
        $c_k$&&&&$\times$\\ \hline
\end{tabular}}
};
\node[] at (0,0.0){
\scalebox{0.555}{
\begin{tabular}{|c||c|c|c|c|}
        \hline
        \color{red} $\mathbb{O}_{\mathbb{D}}(\text{NN})$ &$n_1$&$n_2$& \dots &$n_h$\\ \hline \hline
        $o_1$&$\times$&&&$\times$\\ \hline
        $o_2$&$\times$&&&\\ \hline
        \dots&&\phantom{$\times$}&&\\ \hline
        $o_t$&&&&$\times$\\ \hline
\end{tabular}}
};
\node[] at (0,-1.6){
\hspace{-0.38cm}
\scalebox{0.56}{
  \begin{cxt} 
    \cxtName{$\Scon_{N}$}
          \att{$S_{m_1}$}
          \att{$S_{m_2}$}
          \att{\dots}
          \att{$S_{m_l}$}
          \obj{xx.x}{$n_1$}
          \obj{...x}{$n_2$}
          \obj{....}{\dots}
          \obj{x..x}{$n_h$}
          \end{cxt}}
};
\end{scope}
    \end{tikzpicture}

%% file: pics/fnn.tikz
\begin{tikzpicture}[x=1.5cm, y=1.5cm, >=stealth,scale=0.4,transform shape]
\tikzset{%
  every neuron/.style={
    circle,
    draw,
    minimum size=1cm
  },
  neuron missing/.style={
    draw=none, 
    scale=4,
    text height=0.333cm,
    execute at begin node=\color{black}$\vdots$
  },
}

\foreach \m/\l [count=\y] in {1,2,3,missing,4}
  \node [every neuron/.try, neuron \m/.try] (input-\m) at (0,2.5-\y) {};

\foreach \m [count=\y] in {1,missing,2}
  \node [every neuron/.try, neuron \m/.try ] (mid-\m) at (1.5,2-\y*1.25) {};

\foreach \m [count=\y] in {1,missing,2}
  \node [every neuron/.try, neuron \m/.try ] (hidden-\m) at (3,2-\y*1.25) {};

\foreach \m [count=\y] in {1,missing,2}
  \node [every neuron/.try, neuron \m/.try ] (output-\m) at (4.5,1.5-\y) {};

\foreach \l [count=\i] in {1,2,3,m}
  \draw [<-] (input-\i) -- ++(-1,0)
    node [above, midway] {$v_\l$};

\foreach \l [count=\i] in {1,h}
  \node [above] at (hidden-\i.north) {$n_\l$};%

\foreach \l [count=\i] in {1,k}
  \draw [->] (output-\i) -- ++(1,0)
    node [above, midway] {$c_\l$};


\foreach \i in {1,...,4}
  \foreach \j in {1,...,2}
    \draw [->] (input-\i) -- (mid-\j);

\foreach \i in {1,...,2}
  \foreach \j in {1,...,2}
    \draw [->] (mid-\i) -- (hidden-\j);

\foreach \n [count=\i] in {1,h}
  \foreach \c [count=\j] in {1,k}
    \draw [->] (hidden-\i) -- (output-\j) node[pos=0.8,above] {$w_{\c,\n}$};

\node [align=center, above] at (0,2) {Input \\ layer};        
\node [align=center, above] at (2,2) {Hidden \\ layer};        
\node [align=center, above] at (4,2) {Output \\ layer};        
\end{tikzpicture}

%% file: pics/concepts/vggplum.tikz
\colorlet{mivertexcolor}{blue!50}
\colorlet{jivertexcolor}{orange!80}
\colorlet{vertexcolor}{white}
\colorlet{bordercolor}{white}
\colorlet{linecolor}{gray}
\tikzset{vertexbase/.style 2 args={semithick, shape=circle split, inner sep=5pt, outer sep=0pt, draw=bordercolor},%
  vertex/.style 2 args={vertexbase={#1}{}, shape=circle, draw=linecolor},%
  mivertex/.style 2 args={vertexbase={#1}{}, circle split part fill={vertexcolor, mivertexcolor}},%
  jivertex/.style 2 args={vertexbase={#1}{}, circle split part fill={jivertexcolor, vertexcolor}},%
  divertex/.style 2 args={vertexbase={#1}{}, circle split part fill={jivertexcolor, mivertexcolor}},%
  conn/.style={-, thick, color=linecolor}%
}
\tikzstyle{v} = [label distance =0.3cm,text=red!80!black]
\tikzstyle{a} = [label distance =0.3cm]
\tikzstyle{o} = [text width=3.5cm,label distance=0.2cm]
\tikzstyle{ol} = [text width=3.5cm,label distance=0.2cm,xshift=-1cm]
\tikzstyle{or} = [text width=3.5cm,label distance=0.2cm,xshift=1cm]
\begin{tikzpicture}
  \begin{scope} 
    \begin{scope} 
      \foreach \nodename/\nodetype/\param/\xpos/\ypos in {%
        0/vertex//14/1,
        1/vertex//17.142324121178792/4.35535874696366,
        2/vertex//11.218403914019891/4.440595008937166,
        3/jivertex//14.031200559145699/4.568449401897432,
        4/jivertex//17.227560383152298/7.764809225904031,
        5/vertex//14.159054952105965/7.850045487877541,
        6/jivertex//11.090549521059625/7.93528174985105,
        7/vertex//25.793804711489987/8.787644369586143,
        8/vertex//3/10,
        9/jivertex//14.201673083092718/11.813531669645727,
        10/jivertex//26.00689536642376/12.154476717539765,
        11/vertex//23.023626197350932/12.239712979513271,
        12/vertex//28.73445574957606/12.282331110500028,
        13/vertex//6/13,
        14/mivertex//0.222926119437183/13.049457468261611,
        15/jivertex//2.822632109629218/13.262548123195383,
        16/jivertex//23.066244328337692/15.649163458453646,
        17/jivertex//28.649219487602554/15.734399720427156,
        18/vertex//26.00689536642376/15.777017851413913,
        19/divertex//0.222926119437183/16.927707388056287,
        20/mivertex//3/17,
        21/jivertex//6/17,
        22/vertex//14/17,
        23/jivertex//26.04951349741052/19.527413378248323,
        24/divertex//3.20619528851001/20.678102914890697,
        25/jivertex//14/21,
        26/vertex//17/21,
        27/mivertex//10.8348407351391/21.019047962784736,
        28/divertex//11/25,
        29/mivertex//14/25,
        30/jivertex//17/25,
        31/divertex//14/30 
      } \node[\nodetype={\param}{}] (\nodename) at (\xpos, \ypos) {};
    \end{scope}
    \begin{scope} 
      \path (4) edge[conn] (9);
      \path (6) edge[conn] (19);
      \path (26) edge[conn] (29);
      \path (5) edge[conn] (9);
      \path (10) edge[conn] (25);
      \path (3) edge[conn] (10);
      \path (1) edge[conn] (4);
      \path (14) edge[conn] (20);
      \path (25) edge[conn] (30);
      \path (7) edge[conn] (22);
      \path (14) edge[conn] (19);
      \path (30) edge[conn] (31);
      \path (0) edge[conn] (8);
      \path (20) edge[conn] (29);
      \path (1) edge[conn] (13);
      \path (26) edge[conn] (30);
      \path (27) edge[conn] (28);
      \path (7) edge[conn] (11);
      \path (8) edge[conn] (14);
      \path (20) edge[conn] (24);
      \path (21) edge[conn] (24);
      \path (14) edge[conn] (27);
      \path (8) edge[conn] (13);
      \path (16) edge[conn] (28);
      \path (12) edge[conn] (26);
      \path (5) edge[conn] (18);
      \path (6) edge[conn] (16);
      \path (0) edge[conn] (3);
      \path (2) edge[conn] (14);
      \path (2) edge[conn] (5);
      \path (17) edge[conn] (30);
      \path (19) edge[conn] (24);
      \path (13) edge[conn] (26);
      \path (3) edge[conn] (4);
      \path (22) edge[conn] (27);
      \path (12) edge[conn] (17);
      \path (15) edge[conn] (21);
      \path (25) edge[conn] (28);
      \path (11) edge[conn] (27);
      \path (23) edge[conn] (31);
      \path (24) edge[conn] (31);
      \path (21) edge[conn] (30);
      \path (3) edge[conn] (15);
      \path (4) edge[conn] (17);
      \path (19) edge[conn] (28);
      \path (10) edge[conn] (16);
      \path (13) edge[conn] (21);
      \path (0) edge[conn] (1);
      \path (9) edge[conn] (23);
      \path (18) edge[conn] (23);
      \path (22) edge[conn] (26);
      \path (17) edge[conn] (23);
      \path (4) edge[conn] (21);
      \path (2) edge[conn] (6);
      \path (5) edge[conn] (20);
      \path (13) edge[conn] (20);
      \path (15) edge[conn] (25);
      \path (8) edge[conn] (15);
      \path (8) edge[conn] (22);
      \path (6) edge[conn] (9);
      \path (18) edge[conn] (29);
      \path (7) edge[conn] (12);
      \path (28) edge[conn] (31);
      \path (9) edge[conn] (24);
      \path (15) edge[conn] (19);
      \path (10) edge[conn] (17);
      \path (11) edge[conn] (18);
      \path (1) edge[conn] (12);
      \path (2) edge[conn] (11);
      \path (7) edge[conn] (10);
      \path (29) edge[conn] (31);
      \path (11) edge[conn] (16);
      \path (3) edge[conn] (6);
      \path (27) edge[conn] (29);
      \path (12) edge[conn] (18);
      \path (0) edge[conn] (7);
      \path (16) edge[conn] (23);
      \path (1) edge[conn] (5);
      \path (0) edge[conn] (2);
      \path (22) edge[conn] (25);
    \end{scope}
    \begin{scope} 
      \foreach \nodename/\labelpos/\labelopts/\labelcontent in {%
        23/above/a/{+11},
        24/above/a/{+2},
        28/above/a/{+4},
        29/above/a/{+6},
        30/above/a/{+0},
        0/below/o/{Plum, Plum 2, Corn, Huckleberry, Cucumber Ripe, Mandarine, Salak},
        1/below/or/{Strawberry Wedge},
        2/below/ol/{Onion Red Peeled, Passion Fruit, Fig, Physalis with Husk, Nectarine Flat, Pear Abate, Beetroot},
        3/below/o/{Clementine, Cauliflower, Apple Pink Lady, Apple Red 1, Pomegranate, Apple Crimson Snow},
        4/below/or/{Pitahaya Red, Apple Braeburn, Pineapple Mini},
        5/below/o/{Tomato not Ripened, Pear Forelle, Onion Red, Pear, Chestnut, Pear Stone},
        6/below/ol/{Pear Red, Nut Forest},
        7/below/o/{Tomato Maroon, Mangostan, Tomato Heart, Cocos},
        8/below/o/{Cantaloupe 2},
        9/below/o/{Granadilla, Mango Red, Strawberry, Nut Pecan, Watermelon, Hazelnut},
        10/below/o/{Physalis, Tangelo, Tamarillo},
        11/below/ol/{Cucumber Ripe 2, Pepper Green},
        12/below/or/{Apple Red Delicious, Kumquats, Cherry 2, Cherry Rainier},
        13/below/or/{Pear Kaiser},
        14/below/ol/{Apple Golden 3, Cherry 1, Avocado, Apple Red 3},
        15/below/o/{Apple Red Yellow 1, Cantaloupe 1, Quince, Lemon Meyer, Pomelo Sweetie, Carambula},
        16/below/ol/{Grapefruit Pink, Potato Red Washed, Kohlrabi},
        17/below/or/{Redcurrant, Raspberry, Mulberry},
        18/below/o/{Pear 2, Melon Piel de Sapo, Cactus fruit},
        19/below/ol/{Kiwi, Papaya, Dates, Peach Flat, Ginger Root, Peach, Orange, Banana, Banana Red},
        20/below/o/{Grape White 4, Blueberry, Guava, Avocado ripe},
        21/below/or/{Pineapple, Nectarine, Pear Monster},
        22/below/o/{Banana Lady Finger, Tomato 2, Pepper Orange, Tomato 3},
        23/below/o/{Limes, Mango, Grape White},
        24/below/o/{Apricot, Walnut, Lemon, Potato Sweet},
        25/below/o/{Tomato 1, Tomato Yellow, Pepper Yellow, Maracuja, Tomato 4, Peach 2},
        26/below/or/{Cherry Wax Yellow, Apple Golden 2, Cherry Wax Red, Rambutan},
        27/below/ol/{Corn Husk, Eggplant, Grape Pink, Apple Granny Smith},
        28/below/ol/{Apple Red 2, Pepino, Tomato Cherry Red},
        29/below/o/{Potato White, Potato Red, Plum 3, Grape Blue, Grape White 2, Onion White, Grape White 3, Cherry Wax Black, Pear Williams},
        30/below/or/{Pepper Red, Lychee},
        31/below/o/{Grapefruit White, Apple Red Yellow 2, Apple Golden 1, Kaki},
        0/right/v/{7},
        1/right/v/{8},
        2/right/v/{14},
        3/right/v/{13},
        4/right/v/{17},
        5/right/v/{21},
        6/right/v/{22},
        7/right/v/{11},
        8/right/v/{8},
        9/right/v/{38},
        10/right/v/{20},
        11/right/v/{20},
        12/right/v/{16},
        13/right/v/{10},
        14/right/v/{19},
        15/right/v/{20},
        16/right/v/{34},
        17/right/v/{31},
        18/right/v/{34},
        19/right/v/{42},
        20/right/v/{31},
        21/right/v/{28},
        22/right/v/{16},
        23/right/v/{63},
        24/right/v/{70},
        25/right/v/{37},
        26/right/v/{26},
        27/right/v/{33},
        28/right/v/{71},
        29/right/v/{65},
        30/right/v/{58},
        31/right/v/{131}
      } \coordinate[label={[\labelopts]\labelpos:{\labelcontent}}](c) at (\nodename);
    \end{scope}
  \end{scope}
\end{tikzpicture}